\documentclass[lettersize,journal]{IEEEtran}
\usepackage{epsfig} % for postscript graphics files
\usepackage{subcaption}
\usepackage{graphicx}
\usepackage{cite}
\usepackage{amsmath,amsfonts,amssymb,bm}
\usepackage{mathtools,nccmath}
\usepackage{xcolor}
\usepackage[ruled,vlined]{algorithm2e}
\usepackage{url}
\usepackage{xcolor}
\UseRawInputEncoding
\begin{document}

\title{Bayesian Heuristics for Robust Spatial Perception}

\author{Aamir Hussain Chughtai, Muhammad Tahir and Momin Uppal% <-this % stops a space
	%\thanks{*This work was not supported by any organization}% <-this % stops a space
	\thanks{The authors are with Department of Electrical Engineering, Lahore University of Management Sciences, DHA Lahore Cantt., 54792, Lahore Pakistan. (email: chughtaiah@gmail.com; tahir@lums.edu.pk; momin.uppal@lums.edu.pk) }
	%        {\tt\small albert.author@papercept.net}}%
%\thanks{$^{2}$Bernard D. Researcheris with the Department of Electrical Engineering, Wright State University,
	%        Dayton, OH 45435, USA
	%        {\tt\small b.d.researcher@ieee.org}}%
}


\maketitle

\begin{abstract}
Spatial perception is a key task in several machine intelligence applications such as robotics and computer vision. In general, it involves the nonlinear estimation of hidden variables that represent the system's state. However, in the presence of measurement outliers, the standard nonlinear least squared formulation results in poor estimates. Several methods have been considered in the literature to improve the reliability of the estimation process. Most methods are based on heuristics since guaranteed global robust estimation is not generally practical due to high computational costs. Recently general purpose robust estimation heuristics have been proposed that leverage existing non-minimal solvers available for the outlier-free formulations without the need for an initial guess. In this work, we propose three Bayesian heuristics that have similar structures. We evaluate these heuristics in practical scenarios to demonstrate their merits in different applications including 3D point cloud registration, mesh registration and pose graph optimization. The general computational advantages our proposals offer make them attractive candidates for spatial perception tasks.   
\end{abstract}

\begin{IEEEkeywords}
Spatial Perception, Measurement Outliers, Nonlinear Estimation, Variational Bayes, Expectation-Maximization, Statistical Inference, Parameter and State Estimation.
\end{IEEEkeywords}

\section{Introduction}
Several machine intelligence tasks in robotics depend on reliable spatial perception which involves the estimation of the unknown latent variable representing the state describing the system. Examples of spatial perception include object detection and localization, motion estimation, simultaneous localization and mapping (SLAM) \cite{5540108,6096039,7747236,10147858} etc. The information for inference is available in form of noisy observations which can generally be represented as transformations of the hidden variable given as
\begin{equation}
	\mathbf{y}_i = \mathbf{h}_i(\mathbf{x})+\mathbf{\epsilon}_i \label{eq1}
\end{equation}
where the \textit{i}th measurement $\mathbf{y}_i$ ($i=1,\hdots,m$), of a batch of data $\mathbf{y}$, is expressed as a known nonlinear function $\mathbf{h}_i(.)$ of the unknown variable of interest $\mathbf{x}$ corrupted by random noise $\mathbf{\epsilon}_i$. Under the assumption that $\mathbf{\epsilon}_i$, described with zero-mean Gaussian noise statistics with the precision matrix $\bm{\Omega}_i$ (inverse of the covariance matrix), is uncorrelated across each measurement channel, the \textit{maximum a posteriori} (MAP) estimate has the following equivalent least square formulation \cite{7747236}
\begin{equation}
	\underset{\mathbf{x} \in \mathcal{X}}{\operatorname{argmin}} \sum_{i=0}^m\left\|\mathbf{y}_i-\mathbf{h}_i(\mathbf{x})\right\|_{\bm{\Omega}_i}^2=\underset{\mathbf{x} \in \mathcal{X}}{\operatorname{argmin}}\sum_{i=0}^m {\big(r_i\left(\mathbf{y}_i, \mathbf{x}\right)\big)}^2 \label{eq2}
\end{equation}
%\begin{equation}
%	\min _{\mathbf{x} \in \mathcal{X}} \underset{\mathbf{x} \in \mathcal{X}}{\operatorname{argmin}} \sum_{i=0}^m {\big(r_i\left(\mathbf{y}_i, \mathbf{x}\right)\big)}^2 \label{eq2}
%\end{equation}
where $\mathcal{X}$ denotes the domain of $\mathbf{x}$, the notation $\|\mathbf{e}\|_{\bm{\Omega}}^2={\mathbf{e}}^{\top} \bm{\Omega} \mathbf{e}$ and $r_i(\mathbf{y}_i, \mathbf{x})$ is the residual error for the \textit{i}th measurement. Note that $r_0(\mathbf{y}_0, \mathbf{x})$ incorporates the regularizing term considering a Gaussian prior for $\mathbf{x}$. It is well known that the cost function in \eqref{eq2} leads to brittle estimates in face of measurement outliers owing to unwanted over-fitting to the corrupted data \cite{9239326}. The observations can be easily plagued with outliers in practice due to sensor failures, environmental factors, or due to erroneous data association by the preprocessing front-end algorithms \cite{9610021,9399509}. This limits the reliability of perception-based tasks and is therefore a dynamic research area in robotics.

Owing to the underlying functional nonlinearities and nonconvexity of the domain, even solving \eqref{eq2} globally for common spatial perception applications can be challenging. However, several estimators have been devised in this regard for various applications including point cloud registration, mesh registration, pose graph optimization \cite{horn1987closed,8100078,rosen2019se} etc. These are commonly termed as \textit{non-minimal} solvers which utilize all the measurements for estimation. On the other hand, \textit{minimal} solvers use the smallest number of observations for estimation \cite{yang2020graduated}. 

The problem of estimation in the presence of outliers becomes more complicated since the standard cost function in \eqref{eq2} is inadequate for this purpose. Therefore, several approaches have been devised in this regard. Most of the methods are heuristics that offer efficient solutions mostly without guarantees \cite{fischler1981random,8968174,Schonberger_2016_CVPR,9361339,yang2020graduated}. Other specialized guaranteed approaches also exist that provide solutions with formal certificates. However, their practicality gets limited due to issues with scalability to large-scale problems as the underlying semidefinite programs (SDP) can have a very high computational budget \cite{8624393,8258864}. Naturally, efficient heuristics become the default choice to enable such applications. In the literature, hybrid approaches have also been advocated where the solutions obtained from heuristics are subsequently evaluated for optimality \cite{9286491}. Since these hybrid approaches have been found to work effectively in practice, efficient heuristics are highly desired.

Different general purpose heuristics for robust spatial perception have been designed based on consensus maximization aiming to maximize observations within a predefined inlier threshold during estimation \cite{9610021}. The famous random sample consensus (RANSAC) approach \cite{fischler1981random} is extensively employed relying on minimal solvers for operation. Realizing the fragility of RANSAC under a high outlier regime and its scalability issues, Adaptive Trimming (APAPT) has been proposed to address these limitations \cite{8968174}. ADAPT adopts non-minimal solvers in its design.   

Another popular approach is M-estimation \cite{ronchetti2009robust} which relies on robust cost functions applied to the residuals for resilience against data corruption. Since these formulations are generally inefficient to solve globally \cite{9610021}, the recently introduced graduated non-convexity (GNC) approach proposes an efficient heuristic \cite{yang2020graduated}. It replaces the underlying non-convex functions with surrogate functions for two common cost functions namely the Geman McClure (GM) and the truncated least square (TLS). Using the Black-Rangarajan duality it casts an equivalent formulation which is then iteratively solved using variable and weight update steps alternatively resulting in GNC-GM and GNC-TLS methods. These are general-purpose robust estimators that employ non-minimal solvers during the variable update step and require no initial guess for operation.

The relevant literature also indicates the use of Bayesian methods for outlier-robust estimation. In \cite{7139840}, a method to tune M-estimators is suggested. It aims to estimate the tuning parameters of the M-estimators within an Expectation Maximization (EM) framework. However, the overall scheme is complicated due to its reliance on adaptive importance sampling. Another Bayesian EM-based method is reported in \cite{9361339} where the parameters of a general robust function are estimated along with the primary variable of interest. The method invokes iteratively reweighted least squares (IRLS) within its EM framework which adds an additional layer of approximation during inference since IRLS performance can be sensitive to initialization. Other Bayesian methods, reported in the filtering context, do not involve M-estimation for inference. For example, methods in \cite{8869835,8398426} propose handling outliers by modifying the standard Gaussian likelihood distributions and subsequently perform inference. However, these algorithms do not treat outliers independently for each measurement channel due to under-parameterization \cite{chughtai2022outlier}. Obviating this limitation, two similar Bayesian methods have been proposed: the recursive outlier-robust (ROR) \cite{6349794} and the selective observations rejecting (SOR) \cite{chughtai2022outlier} method. We build on these two methods for the general nonlinear robust estimation problem with the following contributions. 
\begin{itemize}
	\item We study the limitations of the standard ROR and SOR methods which shows why these standard approaches falter in spatial perception applications. The analysis suggests the need for adapting the hyper-parameters, governing the outlier characteristics, during inference.  
	\item We propose three methodologies to overcome the shortcomings of the standard approaches. Similar to the GNC methods, we are interested in invoking the existing non-minimal estimators. To this end, we consider point estimates for $\mathbf{x}$ and use the EM framework. Since EM can be viewed as a special case of variational Bayes (VB), the adaptation is possible. The proposed approaches are similar to the iterative GNC methods with alternating variable and weight update steps.
	\item We also evaluate the proposals in several experimental scenarios and benchmark them against the GNC methods which are the state-of-the-art general purpose heuristics for robust spatial perception.    
\end{itemize}

The structure of the remaining paper is as follows. In \mbox{Section \ref{Sec_Bayes}}, we motivate the selection of the ROR and SOR methods which we build upon. Moreover, we discuss the inferential tools employed in our proposals. In \mbox{Section \ref{propose_algos}}, we discuss the limitations of the standard approaches and present methodologies to overcome their shortcomings. In Section \ref{Sec_Exp}, we discuss the experimental results. Lastly, Section \ref{Sec_conc} provides concluding remarks along with some future directions. 

\section{Relevant Bayesian methods and tools}\label{Sec_Bayes}
In this section, we first briefly discuss the motivation for choosing the two Bayesian methods: ROR and SOR. Moreover, we provide a short primer on the Bayesian tools that we leverage in our proposals in the upcoming section.   
\subsection{Choice of the Bayesian methods for robust estimation}
Recently, ROR and SOR methods have been successfully applied for devising robust nonlinear filtering techniques with satisfactory performance results. In these methods, estimation of $\mathbf{x}$ in \eqref{eq1} relies on modifying the measurement noise statistics. The choice is motivated by the inability of the nominal Gaussian noise to describe the data in face of outliers. Subsequently, the noise parameters are jointly estimated with $\mathbf{x}$. Bayesian theory offers attractive inferential tools for estimating the state and parameters jointly enabling iterative solutions \cite{cbst_book,sarkka2023bayesian}. We adapt these methods to the general nonlinear estimation context of \eqref{eq1} and \eqref{eq2} where non-minimal solvers for estimation are available for the outlier-free cases. The motivation for choosing these particular methods is twofold. First, the formulations of these filters lend their modification conveniently to the nonlinear problem at hand. Moreover, owing to the modeling simplicity, the choice of the hyperparameters for the noise statistics is intuitive for adaptation to our case.

\subsection{EM as a special case of VB}
For solving the problem in \eqref{eq2}, we aim to use non-minimal solvers, which have been developed and tested for different applications. To that end, we need to cast the ROR and SOR algorithms in a way that existing nonlinear least squared solvers are invoked during inference. These Bayesian methods are devised using VB which leads to distributions for the state and parameters. However, the available solvers generally result in point estimates for the state. Therefore, to enable adoption of these Bayesian approaches for robust spatial perception applications, we adopt the EM method which as shown in the Bayesian literature can be viewed as a special case of the VB algorithm \cite{vsmidl2006variational}. We first present the VB method and then interpret EM method as its special case.
\subsubsection{VB}
Suppose that we are interested in estimating multivariate parameter $\bm{\theta}$ from data $\mathbf{y}$. For tractability, we can resort to the VB algorithm considering the mean-field approximation where the actual posterior is approximated with a factored distribution as \cite{murphy2013machine}
\begin{equation}
	p(\bm{\theta}|\mathbf{y})\approx \prod_{j=1}^{J} q(\bm{\theta}_j ) \label{mean_field}
\end{equation} 
where $J$ partitions of $\bm{\theta}$ are assumed with the $j$th partition given as $\bm{\theta}_j$. The VB marginals can be obtained by minimizing the Kullback-Leibler (KL) divergence between the product approximation and the
true posterior resulting in 
\begin{align}
	q(\bm{\theta}_j )&\propto e^{\big( \big\langle\mathrm{ln} ( p(\bm{\theta} |\mathbf{y}))\big\rangle_{ q(\bm{\theta}_{-j} ) } \big) } \ \forall \ j \label{VB_update}
\end{align} 
where ${ q(\bm{\theta}_{-j} ) }=\prod_{k\neq j} q(\bm{\theta}_k )$ and $\langle.\rangle_{q(\mathbf{{x}})}$ denotes the expectation of the argument with respect to the distribution $q(\mathbf{{x}})$. The VB marginals can be obtained by iteratively invoking \eqref{VB_update} till convergence. 
\subsubsection{EM}
From the Bayesian literature, we know that the EM method can be viewed as a special case of the VB algorithm considering point densities for some of the factored distributions in \eqref{mean_field}. In particular, the factored distributions which are assumed as point masses in \eqref{mean_field} can be written as delta functions
\begin{align}
	q(\bm{\theta}_n)=\delta(\bm{\theta}_n-\hat{\bm{\theta}}_n ) 
\end{align} 
with $n$ denoting the indices where such assumption is taken. Resultingly, we can update the parameter of $q(\bm{\theta}_n)$ using as  
\begin{equation}
	\hat{\bm{\theta}}_n=\underset{\bm{\theta}_n}{\operatorname{argmax}} \big\langle\mathrm{ln} ( p(\bm{\theta} |\mathbf{y}))\big\rangle_{ q(\bm{\theta}_{-n} ) } \forall \ n	\label{M_step}
\end{equation}
The expression \eqref{M_step} is formally known as the M-Step of the EM method. The remaining factored distributions not considered as point masses can be determined using \eqref{VB_update} where the expectation with respect to $q(\bm{\theta}_n)$ would simply result in sampling $\mathrm{ln} ( p(\bm{\theta} |\mathbf{y}) $ at $\hat{\bm{\theta}}_n$. This is formally called as the E-Step in the EM method.

Treating EM as particular case of VB allows us another advantage in addition to leveraging the existing non-minimal point estimators for the system state. It allows us the liberty to treat those parameters with point masses where the expectation evaluation with respect to that parameter would otherwise be unwieldy.

% Also, the delta function can be viewed as the limiting case of the normal distribution with diminishing variance. The reader can refer to the Appendix for details. We use these observations in our modification of the ROR and SOR methods given as follows.   
%
%\subsection*{Deriving ROR and SOR from the EM Algorithm}
%Considering the delta function for the VB marginal distribution of $\mathbf{x}$ for the iterations i.e. $q^{(l)}(\mathbf{x})=
%\delta(\mathbf{x}-\mathbf{x}^{(l)})$, we can arrive at the EM method. The maximization (M) and the expectation (E) steps of the EM are as follows.
%\subsubsection*{\textnormal{M-Step}}
%\begin{equation*}
%	\mathbf{x}^{(l)}=\underset{\mathbf{x} \in \mathcal{X}}{\operatorname{argmax}}\big\langle\ln p(\mathbf{x},\mathbf{w}|\mathbf{y})\big\rangle_{q^{(l-1)}(\mathbf{w})}	
%\end{equation*}
%\subsubsection*{\textnormal{E-Step}}
%\begin{align*}
%	{q^{(l)}(\mathbf{w})}&=p(\mathbf{w}|\mathbf{y},\mathbf{x}^{(l)})\propto\exp\big\langle\ln p(\mathbf{x},\mathbf{w}|\mathbf{y})\big\rangle_{q^{(l)}(\mathbf{x})}	
%\end{align*}
%It can easily be recognized that the in the M-Step the exponent terms of the normal distributions of $\mathbf{x}$ in the VB-based ROR and SOR method are maximized. This leads to the same variable update step in the ROR and SOR methods. Similarly, the E-Step leads to the weights update steps. In particular, since only the point estimates of $\mathbf{x}$ are involved, the expectations of residuals with respect to $\mathbf{x}$ can be replaced by the residuals evaluated at the point estimates. 

\section{Proposed Algorithms}\label{propose_algos}
Having chosen the two particular methods for application in robust perception tasks and having interpreted EM as a special case of VB, we are in a position to present our proposals. In this section, we first present the standard ROR method and discuss its limitations. Based on the analysis, we propose a methodology to overcome the drawbacks. Then we shift our attention to the SOR method. We present the standard SOR technique and explain its drawbacks. Based on the insights drawn, we propose two frameworks to deal with the shortcomings.
\subsection{ROR Methods}

\subsubsection{Standard ROR}

We use the version of the ROR method as originally reported in Section 2.5 of \cite{6349794} where conditionally independent measurements are considered. The ROR method, as originally reported, assumes $\mathbf{y}_i$ (measurement from each channel) to be scalar but it can be a vector in general. Accordingly, the likelihood to robustify \eqref{eq1}  is the multivariate Student-t density. We denote the distribution as ${\mathrm{St(\mathbf{z}_s|\bm{\phi}_s,\mathbf{\Sigma}_s,\eta)}}$ where the random vector $\mathbf{z}_s$ obeys the Student-t density and the parameters include $\bm{\phi}_s$ (mean), $\mathbf{\Sigma}_s$ (scale matrix) and  $\eta$ (degrees of freedom) which controls the kurtosis or heavy-tailedness. Resultingly, we can write the likelihood density as \cite{6349794}
\begin{align}
	p(\mathbf{y}_i|\mathbf{x})&={\mathrm{St}}\big(\mathbf{y}_i|\mathbf{h}_i(\mathbf{x}),\bm{\Omega}_i^{-1},\nu_i \big)=\int p(\mathbf{y}_i|\mathbf{x},\lambda_i)p(\lambda_i)  d\lambda_i \label{eq_st1}
\end{align}
with the conditional likelihood following the multivariate Gaussian density given as
\begin{equation}
	p(\mathbf{y}_i|\mathbf{x},\lambda_i)=\mathcal{N}(\mathbf{y}_i|\mathbf{h}_i(\mathbf{x}),(\lambda_i \bm{\Omega}_i)^{-1})
\end{equation}
where $\mathcal{N}(\mathbf{z}_n|\bm{\phi}_n,\mathbf{\Sigma}_n)$ symbolizes that the random vector $\mathbf{z}_n$ follows the Gaussian distribution parameterized by $\bm{\phi}_n$ (mean) and $\mathbf{\Sigma}_n$ (covariance matrix). $\lambda_i$ in \eqref{eq_st1} obeys the univariate Gamma distribution given as \cite{6349794} 
\begin{equation}
	p(\lambda_i)=\mathcal{G}(\lambda_i|\frac{\nu_i}{2},\frac{\nu_i}{2})
\end{equation}
where $\mathcal{G}({z}_g|a_g,b_g)$ denotes that the random variable $z_g$ follows the Gamma distribution with the shape parameter $a_g$ and the rate parameter $b_g$ \cite{murphy2007conjugate}.  The normalizing constant of the distribution is denoted as
\begin{equation}
	f(a,b)=\frac{b^a}{\Gamma(a)} \label{norm}
\end{equation} 
where ${\Gamma(a)}$ denotes the Gamma function.

Denoting $\bm{\lambda}$ as the vector with ${\lambda_i}$ its $i$th element, we can write the following using the Bayes theorem
\begin{equation}
	p(\mathbf{x},\bm{\lambda}|\mathbf{y})\propto{p(\mathbf{y}|\bm{{\lambda}},\mathbf{x})	p(\mathbf{x})p(\bm{{\lambda}}) } \label{ROR_1}
\end{equation} 

Resultingly, the log-posterior, can be written as
\begin{align}
	&p(\mathbf{x},\bm{\lambda}|\mathbf{y})= \Big\{\sum_{i=1}^m \Big(-0.5 \lambda_i {\big(r_i\left(\mathbf{y}_i, \mathbf{x}\right)\big)}^2 -0.5 \nu_i \lambda_i  \nonumber \\
	& + (0.5(\nu_i+d)-1) \ln( {\lambda}_i)  \Big) -0.5 {\big(r_0\left(\mathbf{y}_0, \mathbf{x}\right)\big)}^2 + constant \Big\} \label{log_ROR}
\end{align}

To proceed further, we seek the following VB factorization of the posterior distribution
\begin{equation}
	p(\mathbf{x},\bm{{\lambda}}|\mathbf{y})\approx q(\mathbf{x}) q(\bm{{{\lambda}}}) \label{ROR_2}
\end{equation}

Based on \eqref{M_step} and \eqref{log_ROR}, the state variable $\hat{\mathbf{x}}$ is estimated using the VB/EM theory as \cite{6349794}
\begin{equation}
	\hat{\mathbf{x}}=\underset{\mathbf{x} \in \mathcal{X}}{\operatorname{argmin}}\sum_{i=0}^m w_{i} {\left(r_i(\mathbf{y}_i, \mathbf{x})\right)}^2 \label{ROR_x}
\end{equation}
where $	w_{i}=\langle{{\lambda}}_i\rangle_{q({{\lambda}}_i)} \forall\ i>0$. We have considered a point estimator for $\mathbf{x}$ i.e. $q(\mathbf{x})=\delta(\mathbf{x}-\hat{\mathbf{x}})$ to utilize existing solvers for spatial perception tasks mainly available in this form. 

Similarly, the weights can be updated as \cite{6349794}
\begin{equation}
	w_{i}=\left({1+{\frac{{\hat{r}_i^{2}}-d}{\nu+d} }}\right)^{-1}\  \forall\ i>0 \label{ROR_w}
\end{equation}
where $d$ denotes the dimension of $\mathbf{y}_i$ and ${\hat{r}_i^{2}}={\big(r_i\left(\mathbf{y}_i, \hat{\mathbf{x}} \right)\big)}^2$. We assume $\nu_i=\nu\ \forall\ i$ and use $w_i$ to weight the precision matrix instead of $\bar{\lambda}$ as in the original work to remain consistent with comparative works. Since we assume outliers occur only in the measurements, the weight for the regularizing term remains fixed as 1 i.e. $w_0=1$ in \eqref{ROR_x}.

%The ROR method is presented as Algorithm \ref{Algo1}. 
%Note that in the ROR method reported here the weights have been normalized within unity to ensure no discrepancy with the prior weights. 

\subsection*{Limitations of standard ROR} Starting with $w_{i}=1\ \forall\ i$, the standard ROR invokes \eqref{ROR_x} and \eqref{ROR_w} iteratively till convergence. The technique has shown good performance in filtering context for different practical examples such as target tracking \cite{6349794} and indoor localization \cite{chughtai2022outlier}. This can be attributed to the appearance of well-regularized cost functions \eqref{eq2}. However, for advanced problems in robust spatial perception, the performance of standard ROR compromises at high outlier ratios since it fails to capture the outlier characteristics by fixing $\nu$ which governs the heavy-tailedness of noise or the characteristics of outliers. Empirical evidence confirms this observation ({available as supplementary results \cite{git_ref}}). 
\begin{figure}[h!]
	\centering
	\includegraphics[width=.6\linewidth]{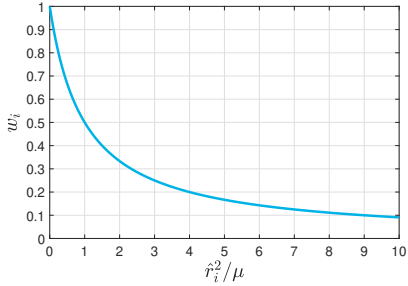}
	\caption{$w_i$ vs ${\hat{r}_{i}^{2}}/\mu$ in ROR.}
	\label{fig:rorweights}
\end{figure}

For further understanding of this limitation consider how $w_i$ changes with parameters of Student-t distribution. The variation of $w_i$ against ${\hat{r}_{i}^{2}}/\mu$ is shown in Fig.~\ref{fig:rorweights} where  $\mu=\nu+d$. The plot of \eqref{ROR_w} would only be a shifted version of the plot in Fig.~\ref{fig:rorweights}. Since $\hat{r}_i^{2} \gg d$ when outlier appears in the \textit{i}th dimension, we can simplify \eqref{ROR_w} as $w_{i}=({1+{\frac{{\hat{r}_i^{2}}}{\mu} }})^{-1} $ indicating the importance of the kurtosis $\nu$ and resultingly $\mu$. It can be observed that the residuals are gradually downweighted with increasing magnitude during estimation with $w_i=0.5$ for ${\hat{r}_{i}^{2}}=\mu$. Since the residuals are evaluated considering the state estimate using all the measurements initially, it is possible that the squared residuals even for the uncorrupted dimensions become greater than the prefixed $\mu$ downweighting them in the process. This can lead to performance issues. Therefore, this calls for adapting $\mu$ by considering the residuals evaluated with clean and corrupted measurements. 
Given the limitations, we propose a variant of the standard ROR method where $\mu$ is adapted during iterations considering the residuals at each iteration. We call it Extended ROR or simply EROR.

\subsubsection{EROR}
%\subsubsection*{Choice of the hyperparameter $\nu$}

%The choice of the hyperparameter $\nu$ is important in the ROR method. The original text \cite{6349794} proposes a constant value of $\nu$ which works generally well for filtering problems due to the availability of the prior estimates of $\mathbf{x}$. However, for general estimation problems, especially without the availability of prior for $\mathbf{x}$, a careful selection of $\nu$ is required. 
In EROR, we propose adaptation of $\mu$ during iterations considering the updated squared residuals based on the relationship of the $w_i$ and $\mu$. The choice of $\mu$ is done such that weights assigned to residuals span the maximum portion of the zero to one range. In other words, the largest residuals need to be pruned with the smallest weights in the weighted least squared cost function and vice versa. In particular, for ${{\hat{r}_i^{2}}}={\hat{r}_{\max}^{2}}$ with $w_i\rightarrow0$ leads to $\mu\rightarrow0$ ($\mu=\frac{{\hat{r}_{i}^{2}}}{1/{w_i}-1}$). Practically, $\mu \ll {\hat{r}_{\max}^{2}}$. Similarly for the other extreme ${{\hat{r}_i^{2}}}={\hat{r}_{\min}^{2}}$ with $w_i\rightarrow1$ leads to $\mu\rightarrow\infty$. Practically, $\mu \gg {\hat{r}_{\min}^{2}}$. To cater for both extremes we propose $\mu= \mathrm{mean}({\hat{r}_{\max}^{2}},{\hat{r}_{\min}^{2}})$. Also, $\mu$ is lower bounded by $\chi$ to ensure residuals within a minimum threshold are not neglected during estimation. The notion of $\chi$ is similar to $\bar{c}^2$ as in \cite{yang2020graduated} which is set as the maximum error expected for the inliers. Note that adaptation of $\mu$ is intuitive which can be viewed as an additional step in the standard ROR devised using VB. EROR is presented as Algorithm \ref{Algo1}.

%\begin{algorithm}[h!]
%	\SetAlgoLined
%	Initialize $\text{the iteration index} \ l=0, w_i^{(l)}=1\ \forall\ i  $\;
%	%		Evaluate $\mathbf{x}^{(l)}$ with \eqref{eq2}\;
%	\While{$\textnormal{the convergence criterion has not met}$}{
	%		$l=l+1$\\
	%		Variable update: $\mathbf{x}^{(l)}=\underset{\mathbf{x} \in \mathcal{X}}{\operatorname{argmin}}\sum_{i} w_i^{(l-1)} {(r_i\left(\mathbf{y}_i, \mathbf{x}\right))}^2 $\\
	%		Residual update: ${\hat{r}_i^{2(l)}}={\big(r_i\left(\mathbf{y}_i, \mathbf{x}^{(l)}\right)\big)}^2$ $\forall\ i>0$\\
	%		Hyperparameter update:\ \ \ \ \ \ \ \ \ \
	%		${\hat{r}_{\max}^{2(l)}}= \max_i({\hat{r}_i^{2(l)}}) \ \mathrm{s.t.}\ i>0 $ ${\hat{r}_{\min}^{2(l)}}= \min_i({\hat{r}_i^{2(l)}})\ \mathrm{s.t.}\ i>0 $
	%		$\mu^{(l)}= \mathrm{mean}({\hat{r}_{\max}^{2(l)}},{\hat{r}_{\min}^{2(l)}})$\ \ \ \ \ \ \ \ \ 
	%		$\mu^{(l)}=\max(\mu^{(l)},\chi)$\\		
	%		Weight update: $w_i^{(l)}=\frac{1}{1+({ {\hat{r}_i^{2(l)}} } /{\mu^{(l)}}) }$ $\forall\ i>0$\\
	%	}
%	$\mathbf{x}=\mathbf{x}^{(l)}$
%	\caption{EROR}
%	\label{Algo1}
%\end{algorithm}

\begin{algorithm}[h!]
	\SetAlgoLined
	Initialize $w_i=1\ \forall\ i  $\\
	
	%		Evaluate $\mathbf{x}^{(l)}$ with \eqref{eq2}\;
	\While{$\textnormal{the convergence criterion has not met}$}{
		Variable update: $\hat{\mathbf{x}}=\underset{\mathbf{x} \in \mathcal{X}}{\operatorname{argmin}}\sum_{i} w_i {(r_i\left(\mathbf{y}_i, \mathbf{x}\right))}^2 $\\
		Residual update: ${\hat{r}_i^{2}}={\big(r_i\left(\mathbf{y}_i, \hat{\mathbf{x}} \right)\big)}^2$ $\forall\ i$\\
		Parameteric update:\ \ \ \ \ \ \ \ \ \
		${\hat{r}_{\max}^{2}}= \max_i({\hat{r}_i^{2}}); {\hat{r}_{\min}^{2}}= \min_i({\hat{r}_i^{2}})\ \mathrm{s.t.}\ i>0 $
		%		$\mu= \mathrm{mean}({\hat{r}_{\max}^{2}},{\hat{r}_{\min}^{2}})$\ \ \ \ \ \ \ \ \ 
		$\mu=\max(\mathrm{mean}({\hat{r}_{\max}^{2}},{\hat{r}_{\min}^{2}}),\chi)$\\		
		Weight update: $w_i=\frac{1}{1+({ {\hat{r}_i^{2}} } /{\mu^{}}) }$ $\forall\ i>0$\\
	}
	\caption{EROR}
	\label{Algo1}
\end{algorithm}

%One approach can be to begin with a very large value of $\nu$ (assuming Gaussian measurement noise statistics) and gradually reduce $\nu$ (for heavy-tailed Student-t measurement noise) up to a minimum threshold. This approach works well (tested during simulations) being similar to the GNC method which is based on transition from a convex cost function to a non-convex function for outlier rejection during iterations \cite{yang2020graduated}. 

\subsection{SOR Methods}
\subsubsection{Standard SOR}
The SOR method, as originally reported \cite{chughtai2022outlier}, assumes $\mathbf{y}_i$ to be scalar but it can be a vector in general. In the original work, an indicator vector $\bm{\mathcal{I}}\in\mathbb{R}^m$ with Bernoulli elements is introduced to describe outliers in the measurements. In particular, ${{\mathcal{I}}}_i=\epsilon$ indicates the occurrence of an outlier in the \textit{i}th dimension and ${{\mathcal{I}}}_i=1$ is reserved for the no outlier case. Accordingly, the conditional likelihood to robustify \eqref{eq1} is a multivariate Gaussian density function \cite{chughtai2022outlier}
\begin{align}
	&p(\mathbf{y}_i|\mathbf{x},{\mathcal{I}}_i)={\mathcal{N}}\Big(\mathbf{y}_i|\mathbf{h}_i(\mathbf{x}),({{\mathcal{I}}}_i \bm{\Omega}_i)^{-1} \Big)\nonumber \\
	&=\frac{1}{\sqrt{ {(2 \pi)^{m}{|\bm{\Omega}_i^{-1}|}}  } }  e^{\left( {-}0.5 {{\mathcal{I}}}_i {\left(r_i\left(\mathbf{y}_i, \mathbf{x}\right)\right)}^2 \right)} {{\mathcal{I}_i}}^{0.5} \label{SOR_like}
\end{align}

${\mathcal{I}_i}\ \forall \ i>0$ is assumed to have the following prior distribution
\begin{equation}
	p({{\mathcal{I}}}_i)=(1-{\theta_i}) \delta({{{\mathcal{I}}}_i}-\epsilon)+{\theta_i}\delta( {{{\mathcal{I}}}_i}-1)
\end{equation}
where $\theta_i$ denotes the prior probability of having no outlier in the $i$th measurement channel. $\epsilon$ has the role of catering for describing the anomalous data in effect controlling the covariance of outliers. Using the Bayes theorem we can write 
\begin{equation}
	p(\mathbf{x},\bm{\mathcal{I}}|\mathbf{y})\propto{p(\mathbf{y}|\bm{\mathcal{I}},\mathbf{x})	p(\mathbf{x})p(\bm{\mathcal{I}})}
\end{equation} 
where $\bm{\mathcal{I}}$ denotes the vector with ${\mathcal{I}_i}$ its $i$th element. 

As a result, the log-posterior is given as
\begin{align}
	&\ln (p(\mathbf{x},\bm{\mathcal{I}},b|\mathbf{y}))= \Big\{\sum_{i=1}^m \Big(-0.5 {\mathcal{I}}_i {\big(r_i\left(\mathbf{y}_i, \mathbf{x}\right)\big)}^2 + 0.5 \ln( {\mathcal{I}}_i ) \nonumber \\
	& + \ln\left( (1-{\theta_i})\delta( {{{\mathcal{I}}}_i}-\epsilon)  +{\theta_i}\delta( {{{\mathcal{I}}}_i}-1) \right) \Big) -0.5 {\big(r_0\left(\mathbf{y}_0, \mathbf{x}\right)\big)}^2 \nonumber\\
	&+ constant \Big\} \label{log_ESOR}
\end{align}

For tractable inference we resort to the following VB factorization of the posterior distribution
\begin{equation}
	p(\mathbf{x},\bm{\mathcal{I}}|\mathbf{y})\approx q(\mathbf{x}) q(\bm{{\mathcal{I}}}) \label{VB_SOR}
\end{equation}

Based on \eqref{M_step} and \eqref{log_ESOR}, the state estimate $\hat{\mathbf{x}}$ is updated using the VB/EM theory as \cite{chughtai2022outlier}
\begin{equation}
	\hat{\mathbf{x}}=\underset{\mathbf{x} \in \mathcal{X}}{\operatorname{argmin}}\sum_{i=0}^m w_{i} {(r_i\left(\mathbf{y}_i, \mathbf{x}\right))}^2 \label{SOR_x}
\end{equation}
with $w_{0}=1$ and 
\begin{align}
	w_{i}&=\langle{\mathcal{I}}_i\rangle_{q({\mathcal{I}}_i)}\ \forall\ i>0 \\
	&=\Omega_i+(1-\Omega_i)\epsilon \approx \Omega_i
\end{align}
where $\Omega_i$ parameterizes the VB posterior marginal ${q({\mathcal{I}}_i)}$ corresponding to $\theta_i$ in $p({{\mathcal{I}}}_i)$. $\Omega_i$ is updated using \eqref{VB_update} and \eqref{log_ESOR} as \cite{chughtai2022outlier}

%\begin{equation}
%	w_{i}=\frac{1}{1+{\mathrm{exp}(0.5( {\hat{r}_i^{2}}-\rho^{2} ))}} \forall\ i>0
%\end{equation}

\begin{align}
	\Omega_i &=\frac{1}{1+{\sqrt{\epsilon}}(\frac{1}{\theta_i}-1){ e^ { \left( 0.5 {\hat{r}_i^{2}} (1-\epsilon) \right) } }} 
\end{align}

Since the hyperparameter $\epsilon$ is assumed to be a small positive number we denote it as $\epsilon=\exp({{-{\rho}}^2})$. Assuming a neutral prior for the occurrence of an outlier in the $i$th dimension i.e. ${\theta_i}=0.5$ (as reported originally \cite{chughtai2022outlier}) and noting that $\epsilon=\exp({{-{\rho}}^2}) \approx 0$ we can write 

\begin{equation}
	w_{i}\approx\Omega_i\approx\frac{1}{1+{ e^{\left(0.5({\hat{r}_i^{2}}-\rho^{2} ) \right)} }} \forall\ i>0 \label{SOR_w}
\end{equation}

\subsection*{Limitations of standard SOR} Starting with $w_{i}=1\ \forall\ i$, the standard SOR invokes \eqref{SOR_x} and \eqref{SOR_w} iteratively till convergence. The technique has shown good performance in the filtering context \cite{chughtai2022outlier} but has limited ability in robust spatial perception tasks as the standard ROR method. This drawback compromises the performance especially at high outlier ratios since the standard SOR fails to capture the outlier characteristics by fixing $\rho^2$ (or $\epsilon$) which governs the covariance of outliers. Experimental evidence verifies this limitation ({available as supplementary results \cite{git_ref}}). 

\begin{figure}[h!]
	\centering
	\includegraphics[width=.6\linewidth]{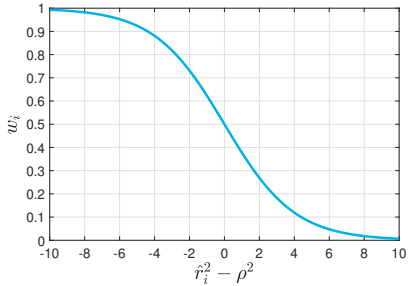}
	\caption{$w_i$ vs ${\hat{r}_{i}^{2}}-\rho^2$ in SOR.}
	\label{fig:sorweights}
\end{figure}

To further appreciate this limitation consider how $w_i$ changes with ${\hat{r}_{i}^{2}}-\rho^2$ as shown in Fig.~\ref{fig:sorweights}. It can be observed that the residuals are gradually downweighted with increasing magnitude during estimation with $w_i=0.5$ for ${\hat{r}_{i}^{2}}=\rho^2$. Since the residuals are evaluated considering the state estimate using all the measurements initially, it is possible that the squared residuals even for the uncorrupted dimensions become greater than the prefixed $\rho^2$ downweighting them during the inference process. This can lead to performance issues. Therefore, this calls for adapting $\rho^2$ by considering the residuals evaluated with clean and corrupted measurements. 

Given the limitations, we present two methods based on the standard SOR technique. Firstly, we propose a modification in the SOR method by explicitly adapting $\rho^2$ during iterations considering the residuals at each iteration. We call it Extended SOR  or simply ESOR. Secondly, by modifying the basic SOR model with a notion of adaptive unknown covariance of outliers, we propose Adaptive SOR or simply ASOR where the parameters controlling the covariance of outliers are \textit{learnt} within the inferential procedure.

%..... it has performance limitations owing to its design. To understand the limitations, we analyze a simple linear example. Consider each measurement, the state variable and its prior to be scalars i.e. $\mathbf{y}_i=y_i$ and $\mathbf{x}=x$ in \eqref{eq2}. Also suppose $\mathbf{h}_i(\mathbf{x})=\mathbf{x}$ and $\bm{\Omega}_i=1\ \forall\ i$. Beginning with \eqref{SOR_x} considering $w_i=1\ \forall\ i>0$, we obtain $\hat{x}=\frac{\sum_i y_i}{m+1}$. In other words the initial estimate of $x$ is the sample mean of all the measurements including those corrupted with outliers. 

\subsubsection{ESOR}
In ESOR, we modify $\rho^2$ during iterations taking into account the updated squared residuals. In particular, we propose to select $\rho^2={\sum_{i=0}^m} w_{i} {\hat{r}_i^{2}}/{\sum_{i=0}^m} w_{i}$. In other words, $\rho^2$ is selected as the the effective centroid of points given as squared residuals considering the weights assigned. With such an intuitive choice, the probability of declaring an outlier in the $i$th dimension is 0.5 when ${\hat{r}_{i}^{2}}$ equals the effective mean of squared residuals. The residuals greater than $\rho$ become the candidates for downweighing and vice versa during an iteration. Lastly, $\rho^2$ is lower bounded by $\gamma$ to ensure residuals within a minimum threshold are not neglected during estimation. ESOR is presented as Algorithm \ref{Algo2}.
%\begin{algorithm}[h!]
%	\SetAlgoLined
%	Initialize $\text{the iteration index} \ l=0, w_i^{(l)}=1\ \forall\ i  $\;
%	%		Evaluate $\mathbf{x}^{(l)}$ with \eqref{eq2}\;
%	\While{$\textnormal{the convergence criterion has not met}$}{
	%		$l=l+1$\\
	%		Variable update: $\mathbf{x}^{(l)}=\underset{\mathbf{x} \in \mathcal{X}}{\operatorname{argmin}}\sum_{i} w_i^{(l-1)} {(r_i\left(\mathbf{y}_i, \mathbf{x}\right))}^2 $\\
	%		Residual update: ${\hat{r}_i^{2(l)}}={\big(r_i\left(\mathbf{y}_i, \mathbf{x}^{(l)}\right)\big)}^2$ $\forall\ i$\\
	%		Hyperparameter update: $\rho^{2(l)}={\sum_i} w_i^{(l-1)} {\hat{r}_i^{2(l)}}/{\sum_i} w_i^{(l-1)}  $\ $\rho^{2(l)}=\max(\rho^{2(l)},\gamma)$\\			
	%		Weight update: $w_i^{(l)}=\frac{1}{1+{\mathrm{exp}(0.5({\hat{r}_i^{2(l)}}-\rho^{2(l)} ))}}$ $\forall\ i>0$\
	%		%			\If{$\underset{i}{\sum} w^{i(l)}=0$}{break\;}
	%		%			Evaluate $\delta={\|\underset{i}{\sum} w^{i(l)} {\hat{r}_i^{2(l)}} -\underset{i}{\sum} w^{i(l-1)} {\hat{r}_i^{2(l-1)}}\|}/{\underset{i}{\sum} w^{i(l-1)} {\hat{r}_i^{2(l-1)}}}$\;
	%	}
%	$\mathbf{x}=\mathbf{x}^{(l)}$\
%	\caption{ESOR}
%	\label{Algo2}
%\end{algorithm}

\begin{algorithm}[h!]
	\SetAlgoLined
	Initialize $w_i=1\ \forall\ i  $\\
	
	%		Evaluate $\mathbf{x}^{(l)}$ with \eqref{eq2}\;
	\While{$\textnormal{the convergence criterion has not met}$}{
		Variable update: $\hat{\mathbf{x}}=\underset{\mathbf{x} \in \mathcal{X}}{\operatorname{argmin}}\sum_{i} w_i {(r_i\left(\mathbf{y}_i, \mathbf{x}\right))}^2 $\\
		Residual update: ${\hat{r}_i^{2}}={\big(r_i\left(\mathbf{y}_i, \hat{\mathbf{x}} \right)\big)}^2$ $\forall\ i$\\
		Parameteric update: $\rho^{2}=\max(\frac{{\sum_i} w_i{\hat{r}_i^{2}}}{{\sum_i} w_i},\gamma)$\\
		Weight update: $w_i=\frac{1}{1+{e^{(0.5({\hat{r}_i^{2}}-\rho^{2} ))} }}$ $\forall\ i>0$\
		%			\If{$\underset{i}{\sum} w^{i(l)}=0$}{break\;}
		%			Evaluate $\delta={\|\underset{i}{\sum} w^{i(l)} {\hat{r}_i^{2(l)}} -\underset{i}{\sum} w^{i(l-1)} {\hat{r}_i^{2(l-1)}}\|}/{\underset{i}{\sum} w^{i(l-1)} {\hat{r}_i^{2(l-1)}}}$\;
	}
	%	$\mathbf{x}=\mathbf{x}^{(l)}$\
	\caption{ESOR}
	\label{Algo2}
\end{algorithm}
%${\hat{r}_i^{2(l)}}={\big(r_i\left(\mathbf{y}_i, \mathbf{x}^{(l)}\right)\big)}^2$ 
%Note that $\theta_i$ is assumed to have a neutral value of 0.5 as specified in the original text. Besides as $\epsilon\approx0$ we ignore $1-\epsilon$ in the exponent for simplicity and assume $\big\langle{\mathcal{I}_i}\big\rangle_{q({\mathcal{I}_i})}=w_i$ since $\epsilon$ has a negligible effect in these calculations.

%\subsubsection*{Choice of the hyperparameter $\rho$}
%
%Setting the hyperparameter $\rho$ properly is important for the performance of the SOR method. Fig.~\ref{fig:sorweights} depicts how $w_i$ changes with ${\hat{r}_{i}^{2}}-\rho^2$. 

%Note that we are working with squared residuals normalized using the inliers precision matrix assumed to be known apriori.

\subsubsection{ASOR}
For devising ESOR we adapt $\rho^2$ (or $\epsilon$) during iterations to capture the characteristics of outliers. Nevertheless, the choice of selection of $\rho^2$ which controls the covariance of outliers is entirely intuitive. This can be viewed as an additional step within the standard SOR method, not falling under the standard VB approach. However, the insights drawn from ESOR with sound experimental performance suggest the merits of considering or \textit{learning} the characteristics of outliers during inference. With these observations, we now present ASOR which is devised with the standard VB approach. In contrast to EROR and ESOR, we jointly \textit{estimate} the covariance controlling factor along with the state and the weights in ASOR.     

The conditional likelihood for designing ASOR remains same as for the standard SOR given in \eqref{SOR_like}. Building on the insights from EROR and ESOR, we need to adapt covariances to describe the outliers. In particular, we assume that for outlier occurrence in the $i$th observation channel i.e. ${{\mathcal{I}}}_i\neq 1$ in \eqref{SOR_like}, ${{\mathcal{I}}}_i$ obeys a Gamma probability density being supported on the set of positive real numbers. Resultingly, we write the hierarchical prior distribution of ${{\mathcal{I}}}_i$ given as
\begin{align}
	p({{\mathcal{I}}}_i|b)&=(1-{\theta_i}) \underbrace{f(a,b) { {{\mathcal{I}}}_i } ^{a-1}  e^{-b {{{\mathcal{I}}}_i} }}_{ \mathcal{G}({{\mathcal{I}}}_i|a,b) } +{\theta_i}\delta( {{{\mathcal{I}}}_i}-1) \label{pIb}
\end{align}
where we assume the two components of $p({{\mathcal{I}}}_i|b)$ in \eqref{pIb} as disjoint with the Gamma density defined as zero for ${{{\mathcal{I}}}_i}=1$ without losing anything. This assumption helps in subsequent derivation. The parameter $b$ is the factor that captures the common effect of outliers in each observation channel. From the Bayesian theory we know that the conjugate prior of $b$ is also a Gamma distribution given as \cite{fink1997compendium} 
\begin{align}
	p(b)&=f(A,B) {b}^{A-1}  e^{-B b } \label{pb_ASOR}
\end{align}
where $A$ and $B$ are the parameters of this Gamma distribution. We are now in a position to invoke the Bayes theorem given as
\begin{equation}
	p(\mathbf{x},\bm{\mathcal{I}},b|\mathbf{y})\propto {p(\mathbf{y}|\bm{\mathcal{I}},\mathbf{x})	p(\mathbf{x})p(\bm{\mathcal{I}}|b)p(b)}
\end{equation} 

Resultingly, the log-posterior, which is used in subsequently derivation, is given as
\begin{align}
	&\ln (p(\mathbf{x},\bm{\mathcal{I}},b|\mathbf{y}))= \Big\{\sum_{i=1}^m \Big(-0.5 {\mathcal{I}}_i {\left(r_i\left(\mathbf{y}_i, \mathbf{x}\right)\right)}^2 + 0.5 \ln( {\mathcal{I}}_i ) \nonumber \\
	& + \ln\left( (1-{\theta_i}) f(a,b) { {{\mathcal{I}}}_i} ^{a-1}  e^{-b {{{\mathcal{I}}}_i} } +{\theta_i}\delta( {{{\mathcal{I}}}_i}-1) \right) \Big) \nonumber \\
	& -0.5 {\big(r_0\left(\mathbf{y}_0, \mathbf{x}\right)\big)}^2 + (A-1)\ln(b)- B b + constant \Big\} \label{log_ASOR}
\end{align}

To proceed further we resort to the VB factorization given as
\begin{equation}
	p(\mathbf{x},\bm{\mathcal{I}},b|\mathbf{y})\approx q(\mathbf{x}) q(\bm{{\mathcal{I}}}) q({b})
\end{equation}

Using the VB/EM theory and with the assumption that $q(\mathbf{x})=\delta(\mathbf{x}-\hat{\mathbf{x}})$ we obtain the following using \eqref{M_step} and \eqref{log_ASOR} 
\begin{equation}
	\hat{\mathbf{x}}=\underset{\mathbf{x} \in \mathcal{X}}{\operatorname{argmin}}\sum_{i=0}^m w_{i} { \left( r_i(\mathbf{y}_i, \mathbf{x}) \right) }^2 
\end{equation}
where $w_0=1$ and
% $	w_{i}=\langle{\mathcal{I}}_i\rangle_{q({\mathcal{I}}_i)} \forall\ i>0$.  	
\begin{align}
	w_{i}&=\langle{\mathcal{I}}_i \rangle_{q({\mathcal{I}}_i)} \ \forall\ i>0  \label{w_i_sora}
\end{align}

Thanks to the notion of VB-conjugacy \cite{vsmidl2006variational}, it turns out that $q(\mathcal{I}_i)$ has a same functional form as of $p({{\mathcal{I}}}_i|b)$ in \eqref{pIb}. $q(\mathcal{I}_i)$ is parameterized by $\alpha$, $\beta_i$ and $\Omega_i$ corresponding to $a$, $b$ and $\theta_i$ in $p({{\mathcal{I}}}_i|b)$ respectively. Resultingly, we can evaluate the expectation in \eqref{w_i_sora} as
\begin{align}
	w_i=\Omega_i+(1-\Omega_i)\alpha/\beta_i \ \forall\ i>0
\end{align}

The VB marginal $q(\bm{\mathcal{I}})$ can be obtained using \eqref{VB_update} and \eqref{log_ASOR} as
\begin{align}
	q(\bm{\mathcal{I}})	 \propto&  \prod_{i=1}^m  \Big\{ \mathcal{I}^{0.5}_i e^{-0.5 {\hat{r}_i^{2} } \mathcal{I}_i} ( (1-{\theta_i}) f(a,b) {{{\mathcal{I}}}_i}^{a-1}  e^{-\hat{b} {{{\mathcal{I}}}_i} } \nonumber \\  &+ {\theta_i}\delta( {{{\mathcal{I}}}_i}-1) ) \Big\} 
\end{align}
where we have assume a point estimator for $b$ i.e. $q(b)=\delta(b-\hat{b})$ to simplify the arising expectations. 

We can further write  
\begin{align}
	q(\bm{\mathcal{I}})	 	 =&  \prod_{i=1}^m  \Big\{    k_i (1-{\theta_i} ) f(a,\hat{b})  {{{\mathcal{I}}}_i}^{\alpha-1 } e^{  - \beta_i \mathcal{I}_i } + \nonumber \\
	&k_i {\theta_i} e^{-0.5 {\hat{r}_i^{2}}}  \delta( {{{\mathcal{I}}}_i}-1) \Big\} \label{k_i} 
\end{align}
where $k_i$ is the proportionality constant for the $i$th dimension, $\beta_i={0.5 {\hat{r}_i^{2}}  +\hat{b} } $ and $\alpha=a+0.5$. 

Proceeding ahead we can write
%\begin{align}
%	q(\bm{\mathcal{I}})	& =  \prod_{i=1}^m    \overbrace{(1-{\Omega_i} )\underbrace{ f(\alpha,\beta_i) {{{\mathcal{I}}}_i}^{\alpha-1 } e^{  - \beta_i \mathcal{I}_i }}_{q^1({\mathcal{I}}_i)}  + \underbrace{{\Omega_i} \delta( {{{\mathcal{I}}}_i}-1)}_{q^2(\mathcal{I}_i)}}^{q(\mathcal{I}_i)} \nonumber   
%\end{align}
\begin{align}
	q(\bm{\mathcal{I}})	& =  \prod_{i=1}^m    \overbrace{(1-{\Omega_i} )\underbrace{ f(\alpha,\beta_i) {{{\mathcal{I}}}_i}^{\alpha-1 } e^{  - \beta_i \mathcal{I}_i }}_{q^1({\mathcal{I}}_i)}  + {{\Omega_i} \delta( {{{\mathcal{I}}}_i}-1)} }^{q(\mathcal{I}_i)} \nonumber   
\end{align}
where 
\begin{equation}
	{\Omega_i}=k_i e^{-0.5 {\hat{r}_i^{2} }} \theta_i \label{omega_sora}
\end{equation}

To determine  $k_i$, we note that the distribution in \eqref{k_i} should integrate to $1$. Therefore, the following should hold
\begin{equation}
	k_i {\theta_i} e^{-0.5 {\hat{r}_i^{2}}}   +   k_i (1-{\theta_i} ) \frac{ f(a,\hat{b})}  {f(\alpha,{\beta_i})  } =1
\end{equation}

Leading to 
\begin{equation}
	k_i=\frac{1}{{\theta_i} e^{-0.5 {\hat{r}_i^{2}}}   +   (1-{\theta_i} ) \frac{ f(a,\hat{b})}  {f(\alpha,{\beta_i})  }} \label{k_sora}
\end{equation}

Resultingly, using \eqref{norm} and \eqref{omega_sora}, \eqref{k_sora} we arrive at 
\begin{equation}
	\Omega_i=\frac{1}{1+ \zeta \frac {{\hat{b}}^a} {{\beta_i^\alpha }} e^{0.5 {\hat{r}_i^{2}} }    } \ \forall\ i>0
\end{equation}
where $\zeta=(\frac{1}{\theta_i}-1) \frac{\Gamma(\alpha)}{\Gamma(a)}$.

Lastly, in a similar manner using the VB/EM approach we can determine $q(b)=\delta(b-\hat{b})$ where using \eqref{M_step} and \eqref{log_ASOR}
\begin{equation}
	\hat{b}=\underset{b}{\operatorname{argmax}} \big\langle\mathrm{ln} ( p(\hat{\mathbf{x}},\bm{\mathcal{I}},b |\mathbf{y})) \big\rangle_{ q(\bm{\mathcal{I}}) } \label{b_ASOR}
\end{equation}

The expected log-posterior in \eqref{b_ASOR} can be written as 
\begin{align}
	\big\langle\mathrm{ln} ( p(\hat{\mathbf{x}},\bm{\mathcal{I}},b |\mathbf{y})) \big\rangle_{ q(\bm{\mathcal{I}}) }=& \Big\{{\sum_{i=1}^m v_i(b)}+{(A-1)}\ln (b){-Bb}\nonumber \\
	&+ constant\Big\} \label{log_post_b}
\end{align}
where
\begin{align}
	v_i(b)& ={   \langle \ln ((1-{\theta_i} ) f(a,b) {{{\mathcal{I}}}_i}^{a-1 } e^{  - b \mathcal{I}_i }  + {\theta_i} \delta( {{{\mathcal{I}}}_i}-1))  \rangle_{q({\mathcal{I}}_i)} }
\end{align}
which can further be written as follows considering only the terms dependent on $b$ 
\begin{align}
	v_i(b) =    (1-\Omega_i) ( \ln ( f(a,b)) - b  \langle{{{\mathcal{I}}}_i}\rangle_{q^1({\mathcal{I}}_i)}) + constant  \label{v_b}
\end{align}
where we assume $q^1(\mathcal{I}_i)$ is defined as zero for ${{{\mathcal{I}}}_i}=1$ as the prior Gamma density in \eqref{pIb}.
%\begin{align}
%	v_i(b) =& \big\{     \Omega_i \ln({\theta_i} ) + (1-\Omega_i) ( \ln ( f(a,b)) + (a-1) \langle{\ln {{\mathcal{I}}}_i}\rangle_{q^1({\mathcal{I}}_i)} \nonumber \\ 
%	&- b \langle{{{\mathcal{I}}}_i}\rangle_{q^1({\mathcal{I}}_i)}  +  \ln ({1-\theta_i})     ) \big\}
%\end{align}
Given $\langle{{{\mathcal{I}}}_i}\rangle_{q^1({\mathcal{I}}_i)}={\alpha}/{\beta_i}$ and using the expressions in \eqref{norm} and \eqref{v_b}, we can write \eqref{log_post_b} as

%\begin{align}
%	\big\langle  \mathrm{ln} ( p(\hat{\mathbf{x}},\bm{\mathcal{I}},b |\mathbf{y})) \big\rangle_{ q(\bm{\mathcal{I}}) }  =& \Big\{  { \sum_{i=1}^m  \big(   a(1-\Omega_i)  \ln  ({b})  - (1-\Omega_i)  \frac{\alpha}{\beta_i} b     \big) }\nonumber \\
%	&+{(A-1)}\ln (b) -Bb + constant\Big\} 
%\end{align}
%which can be written as
\begin{align}
	\big\langle  \mathrm{ln} ( p(\hat{\mathbf{x}},\bm{\mathcal{I}},b |\mathbf{y})) \big\rangle_{ q(\bm{\mathcal{I}}) } =
	&{(\bar{A}-1)}\ln (b) -\bar{B}b + constant \label{lnb_ASOR} 
\end{align}
where
% $\bar{A}= A+\sum_i a (1-\Omega_i) $ and $\bar{B}=B+\sum_i(1-\Omega_i) \frac{\alpha}{\beta_i}$. 
\begin{align}
	\bar{A} &= A+\sum_{i=1}^m a (1-\Omega_i) \label{A_bar}\\
	\bar{B}&=B+\sum_{i=1}^m (1-\Omega_i) \frac{\alpha}{\beta_i} \label{B_bar}
\end{align}

Maximizing \eqref{lnb_ASOR} using differentiation, we obtain $\hat{b}$ according to \eqref{b_ASOR} as
\begin{equation}
	\hat{b}=\frac{\bar{A}-1}{\bar{B}} \ \ \ \mathrm{s.t.} \ \ \ \bar{A}>1 \label{hb_ASOR}
\end{equation}
where $\bar{A}>1$ owing to the requirement of positivity of parameter $b$ of Gamma distribution in \eqref{pIb} being approximated in \eqref{hb_ASOR}. Also noting that the parameter $a>0$ for validity of the Gamma distribution in \eqref{pIb}, $A>1$ is a sufficient condition for \eqref{hb_ASOR} to hold considering \eqref{A_bar}. Lastly, note that since the rate parameter of distribution in \eqref{pb_ASOR} is positive i.e. $B>0$ any numerical errors in \eqref{hb_ASOR} are avoided. 
%Therefore, we can approximate $q(b)$ with a point distribution as $\delta(b-\hat{b})$ where $\hat{b}=\mfrac{\bar{A}-1}{\bar{B}}$. 
The resulting method namely ASOR is given as Algorithm \ref{Algo3}.

\begin{algorithm}[h!]
	\SetAlgoLined
	Initialize $w_i=1\ \forall\ i$\ and $A, B, a, \hat{b}$, $\theta_i\ \forall\ i$ \\
	Evaluate $\alpha=a+0.5$ and $\zeta=(\frac{1}{\theta_i}-1) \frac{\Gamma(\alpha)}{\Gamma(a)}$\\ 
	%		Evaluate $\mathbf{x}^{(l)}$ with \eqref{eq2}\;
	\While{$\textnormal{the convergence criterion has not met}$}{
		Variable update: $\hat{\mathbf{x}}=\underset{\mathbf{x} \in \mathcal{X}}{\operatorname{argmin}}\sum_{i} w_i {(r_i\left(\mathbf{y}_i, \mathbf{x}\right))}^2 $\\
		Residual update: ${\hat{r}_i^{2}}={\big(r_i\left(\mathbf{y}_i, \hat{\mathbf{x}} \right)\big)}^2$ $\forall\ i$\\
		Parameteric updates: \\
		$\beta_i={0.5 {\hat{r}_i^{2}}  +\hat{b} } $ \\
		$	\Omega_i=\frac{1}{1+ \zeta \frac {{\hat{b}}^a} { { \beta_i^\alpha }} e^{0.5 {\hat{r}_i^{2}} } } $ \\
		$\hat{b}=\mfrac{A-1+\sum_i a (1-\Omega_i)}{B+\sum_i(1-\Omega_i) \frac{\alpha}{\beta_i} }$ \\
		Weight update: $ w_i=\Omega_i+(1-\Omega_i)\alpha/\beta_i $  $\forall\ i>0$\ \\
		%			\If{$\underset{i}{\sum} w^{i(l)}=0$}{break\;}
		%			Evaluate $\delta={\|\underset{i}{\sum} w^{i(l)} {\hat{r}_i^{2(l)}} -\underset{i}{\sum} w^{i(l-1)} {\hat{r}_i^{2(l-1)}}\|}/{\underset{i}{\sum} w^{i(l-1)} {\hat{r}_i^{2(l-1)}}}$\;
	}
	%	$\mathbf{x}=\mathbf{x}^{(l)}$\
	\caption{ASOR}
	\label{Algo3}
\end{algorithm}

\subsection*{Remarks}
It is interesting to note that the variable update steps in EROR, ESOR and ASOR is the same as in GNC reflecting their ability to employ non-minimal solvers during inference. We propose using the change in ${\sum_i} w_{i} {\hat{r}_i^{2}}$ during consecutive iterations as the standard convergence criterion. We lastly remark that in EROR and ESOR an exception to break the iterations can be added for seamless operation when the sum of the weights gets close to zero which we experienced in certain applications with very high ratios of outliers.

%Another criterion: $\max_i({w_{i} {\hat{r}_i^{2}}})<\bar{c}^2$ can also be used which obviates the use of $\chi$ and $\gamma$ but can result in deteriorated performance as indicated by .
%The parameters  $\chi$ and $\gamma$ can  easily be set in a way so that weights for residuals greater than $\bar{c}^2$ are assigned much smaller values e.g. $\chi=\gamma=\bar{c}^2/2$ . However, we have obtained acceptable results even by simply setting $\chi$ and $\gamma$ as $\bar{c}^2$ during experiments for different applications under various scenarios.
\section{Experiments}\label{Sec_Exp}
In this section, we discuss the performance results of the proposed methods for different spatial perception applications including 3D point cloud registration (code: MATLAB), mesh registration (code: MATLAB), and pose graph optimization (PGO) (code: C++) on an Intel i7-8550U processor-based computer and consider SI units. We consider GNC-GM and GNC-TLS as the benchmark methods which have shown good results in these applications outperforming methods including the classical RANSAC and recently introduced ADAPT. For EROR and ESOR we consider $\chi=\gamma=\bar{c}^2$ where $\bar{c}^2$, dictating the maximum error expected for the inliers, is set as specified in the original work \cite{yang2020graduated}. In ASOR we resort to the choice of initialization which performs the best across different applications given as $a=0.5,A=10000,B=1000,\hat{b}=10000,\theta_i=0.5\ \forall \ i$. We use the normalized incremental change of $10^{-5}$ in  $\underset{i}{\sum} w^{i} {\hat{r}_i^{2}}$ during consecutive iterations as the convergence criterion for GNC-TLS, EROR, ESOR and ASOR. For GNC-GM the convergence criterion remains the same as originally reported.

\subsection{3D Point Cloud Registration}
\begin{figure}[h!]
	\centering
	\includegraphics[width=.9\linewidth,trim=1cm 3cm 2.5cm 3cm,clip]{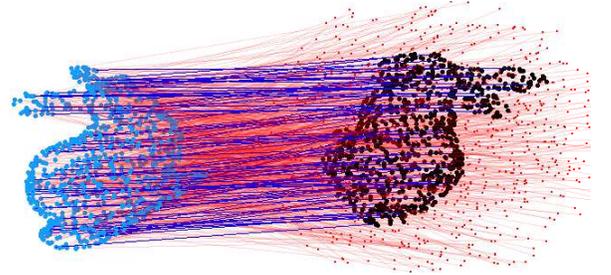}
	\caption{Point clouds with correspondences in 3D point cloud registration for the \textit{Bunny} dataset \cite{curless1996volumetric}.}
	\label{fig:bunny}
\end{figure}
In 3D point cloud registration, we assume that a set of 3D points $\mathbf{p}_i\in \mathbb{R}^3, i = 1, . . . , m$ undergo a transformation, with rotation $\mathbf{R} \in \mathrm{SO(3)}$  and translation $\mathbf{t} \in \mathbb{R}^3$, resulting in another set of 3D points $\mathbf{q}_i\in \mathbb{R}^3, i = 1, . . . , m$. The putative correspondences $(\mathbf{p}_i,\mathbf{q}_i)$ can be potentially infested with outliers. Fig.~\ref{fig:bunny} depicts how the \textit{Bunny} point cloud from the Stanford repository \cite{curless1996volumetric} undergoes a random transformation in a point cloud registration setup (blue lines: inliers, red lines: outliers). The objective is to estimate $\mathbf{R}$ and $\mathbf{t}$ that best aligns the two point clouds by minimizing the effect of outliers. The problem can be cast in form of \eqref{eq2} where the \textit{i}th residual is the Euclidean distance between $\mathbf{q}_i$ and $\mathbf{R}\mathbf{p}_i+\mathbf{t}$. We resort to the renowned Horn's method as the non-minimal solver for this case which provides closed form estimates in the outlier-free case \cite{horn1987closed}.

\begin{figure*}[t!]
	\centering
	\begin{subfigure}{0.32\textwidth}
		\includegraphics[width=\textwidth]{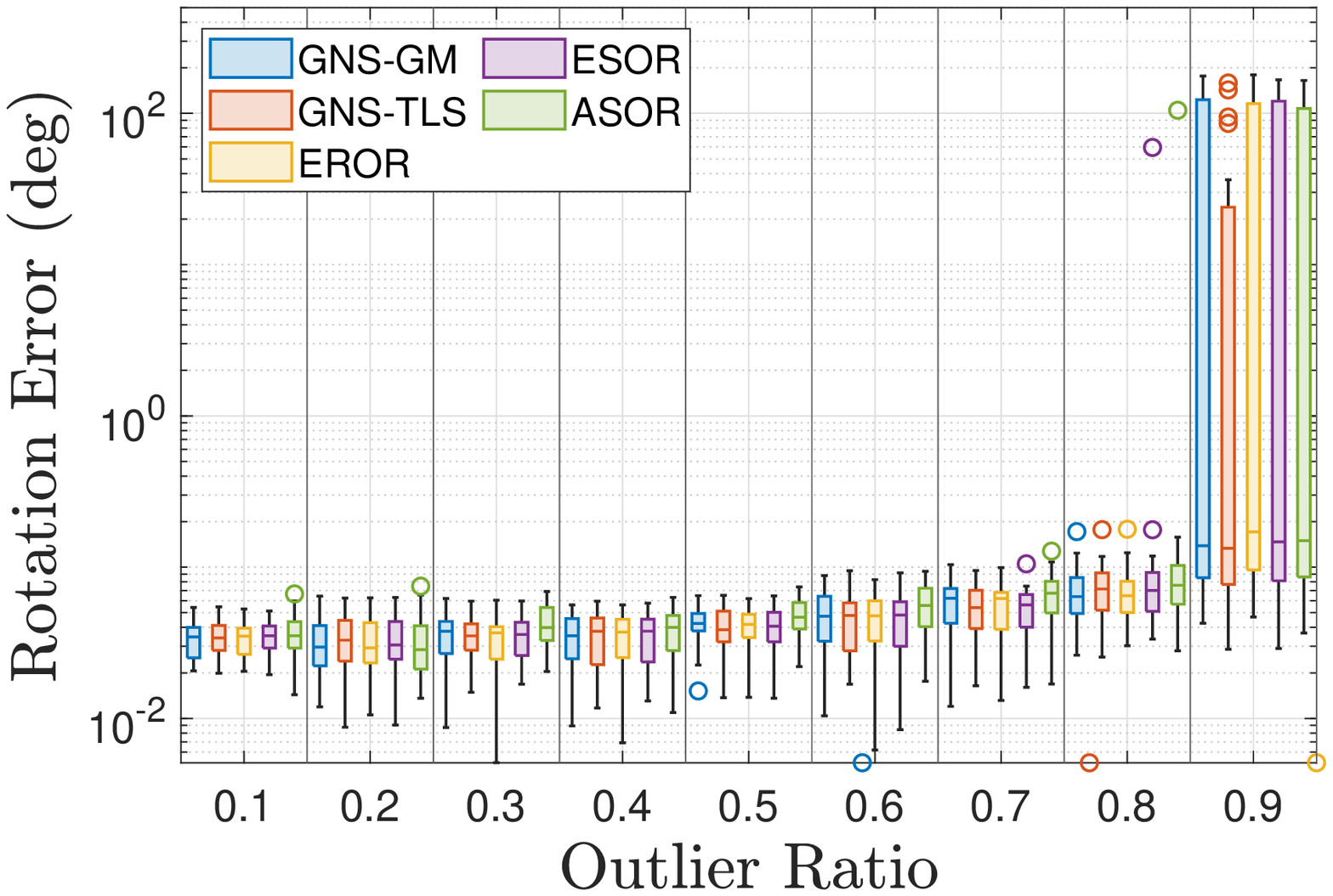}
		\caption{Rotation error.}
		\label{fig:first_horn}
	\end{subfigure}
	\hfill
	\begin{subfigure}{0.32\textwidth}
		\includegraphics[width=\textwidth]{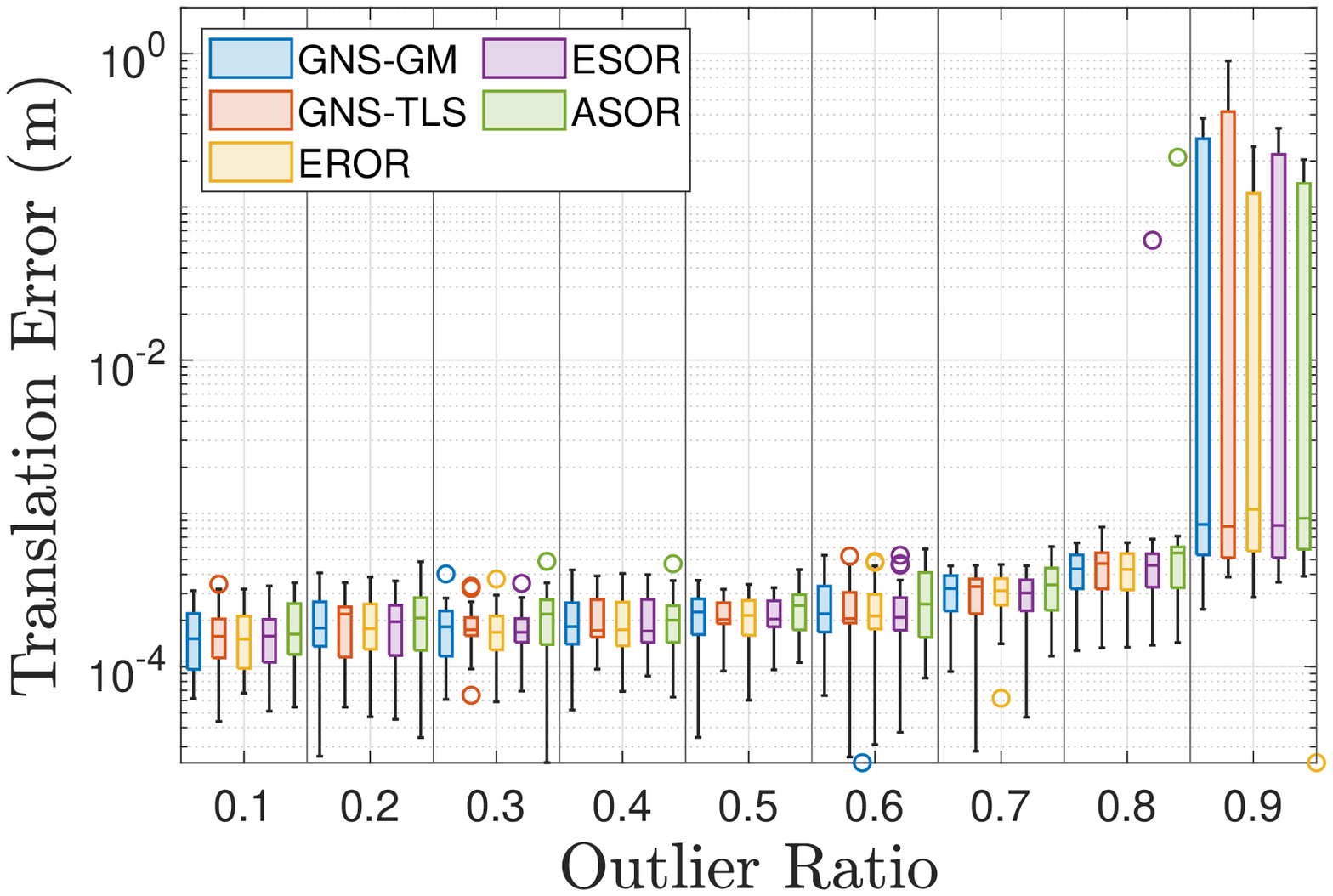}
		\caption{Translation error.}
		\label{fig:second_horn}
	\end{subfigure}
	\hfill
	\begin{subfigure}{0.32\textwidth}
		\includegraphics[width=\textwidth]{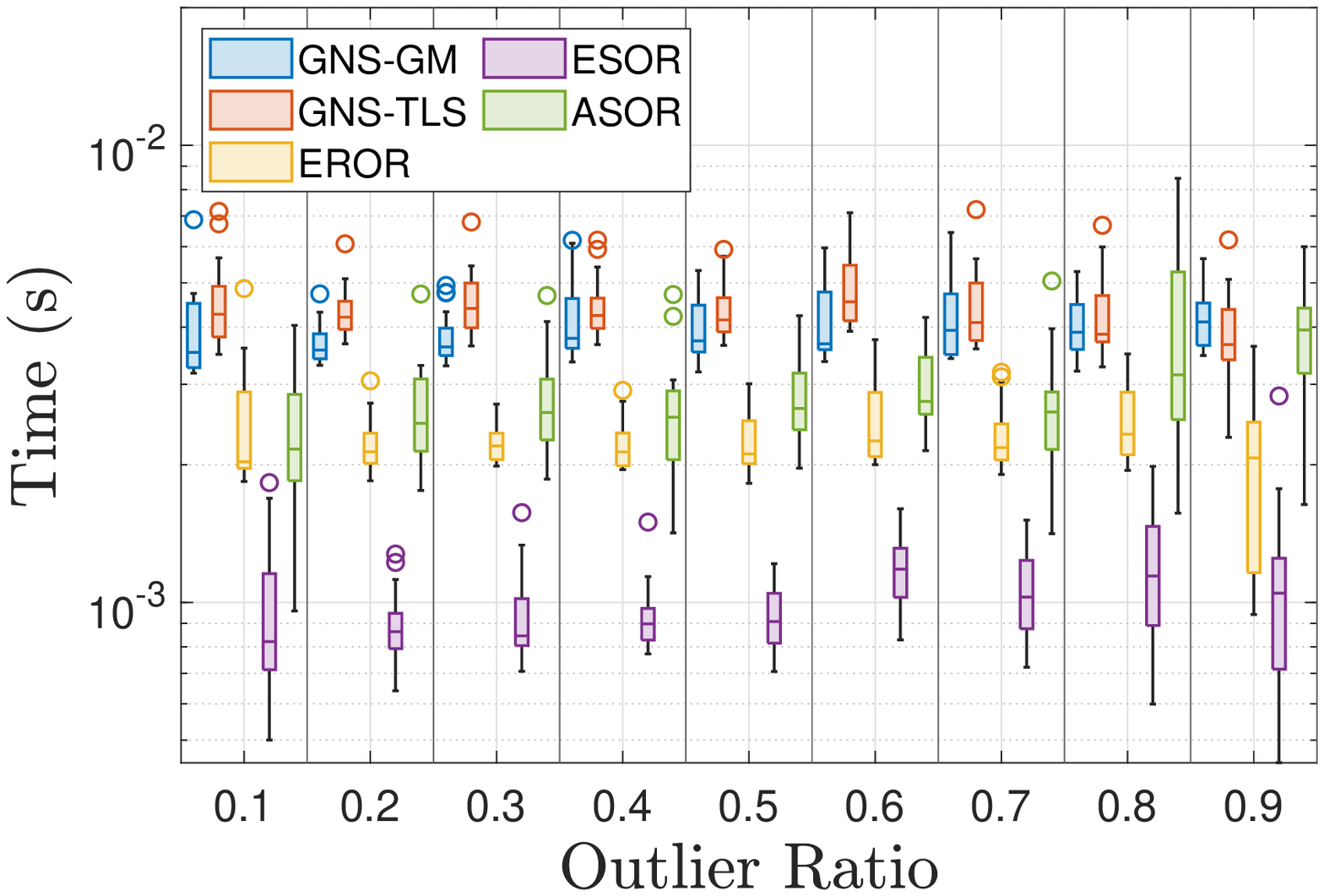}
		\caption{Computational time.}
		\label{fig:third_horn}
	\end{subfigure}
	\caption{Performance of robust estimators for 3D point cloud registration considering the \textit{Bunny} dataset \cite{curless1996volumetric}.}
	\label{fig:figures_horn}
\end{figure*}

\begin{figure*}[t!]
	\centering
	\begin{subfigure}{0.32\textwidth}
		\includegraphics[width=\textwidth]{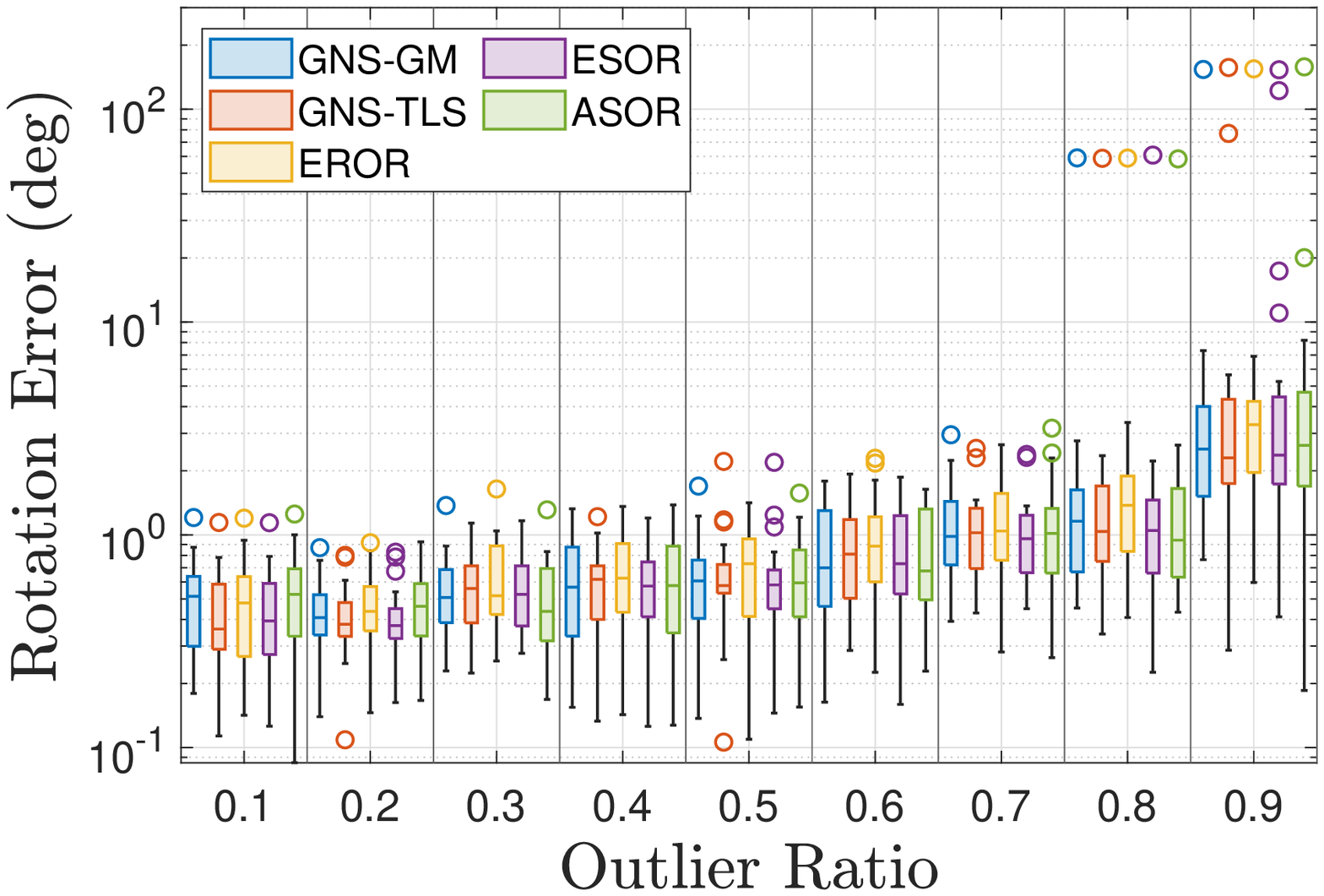}
		\caption{Rotation error.}
		\label{fig:first_mesh}
	\end{subfigure}
	\hfill
	\begin{subfigure}{0.32\textwidth}
		\includegraphics[width=\textwidth]{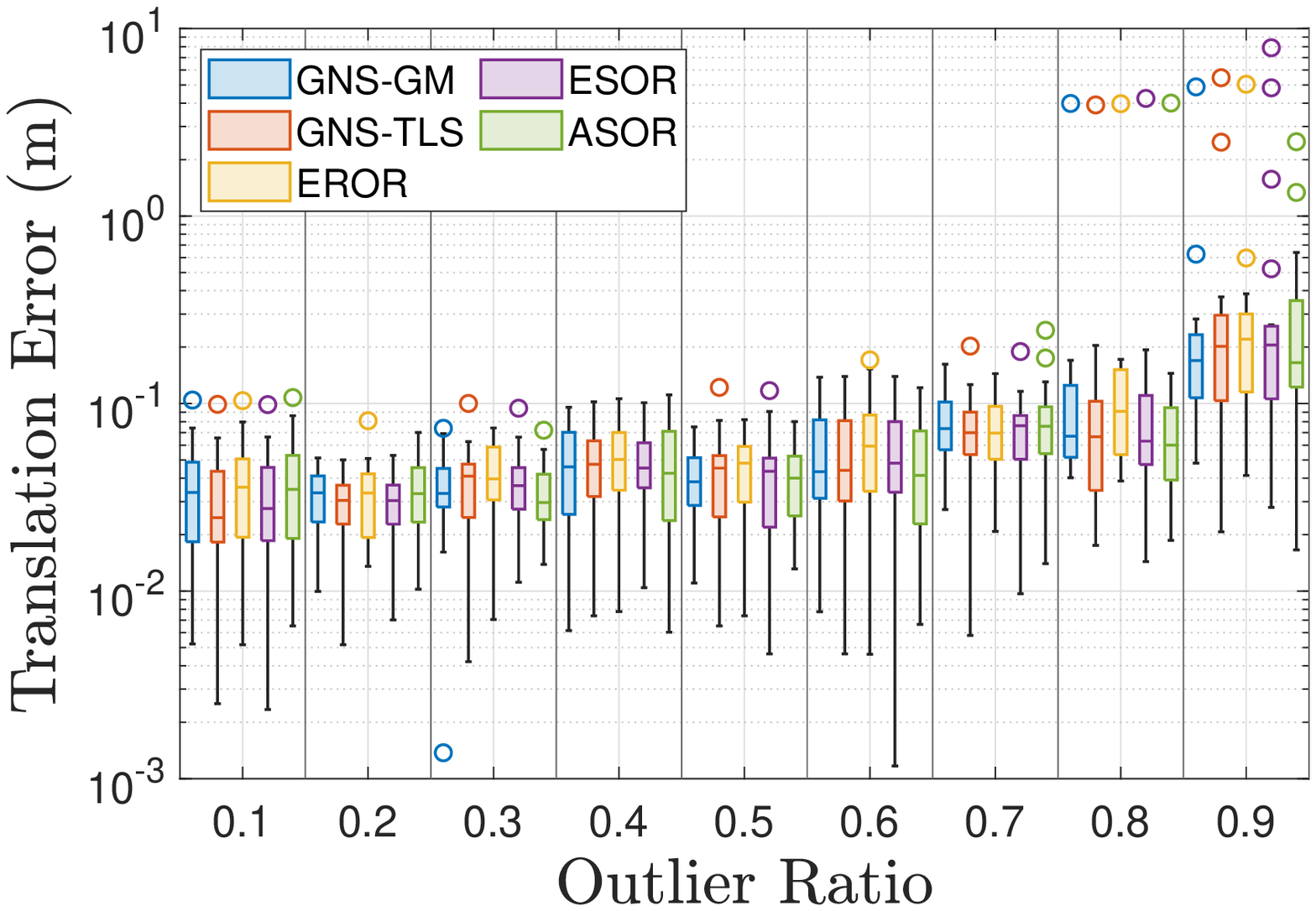}
		\caption{Translation error.}
		\label{fig:second_mesh}
	\end{subfigure}
	\hfill
	\begin{subfigure}{0.32\textwidth}
		\includegraphics[width=\textwidth]{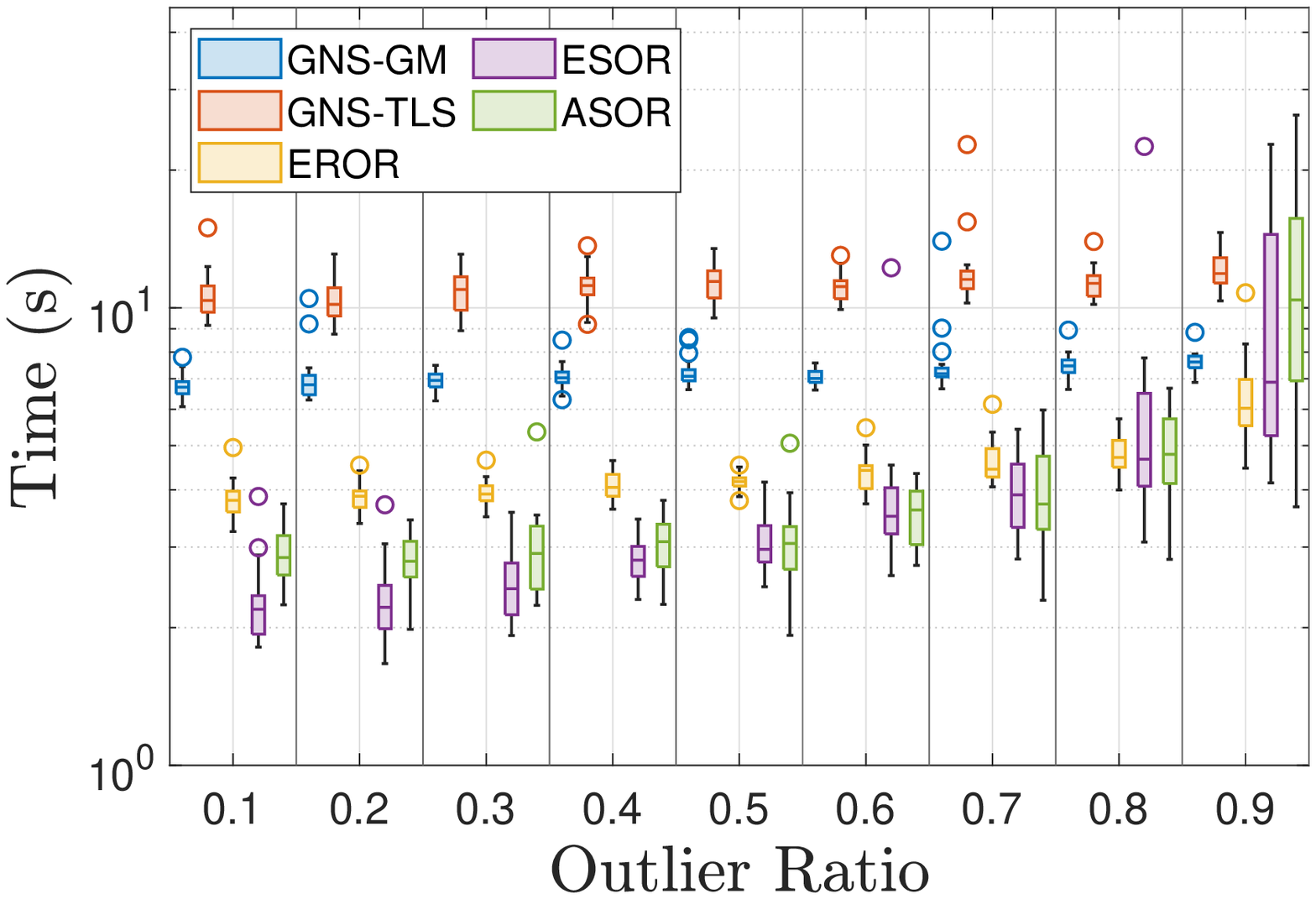}
		\caption{Computational time.}
		\label{fig:third_mesh}
	\end{subfigure}
	\caption{Performance of robust estimators for mesh registration considering the \textit{motorbike} model \cite{6836101}.}
	\label{fig:figures_mesh}
\end{figure*}  

\begin{figure*}[h!]
	\centering
	\begin{subfigure}{0.32\textwidth}
		\includegraphics[width=\textwidth]{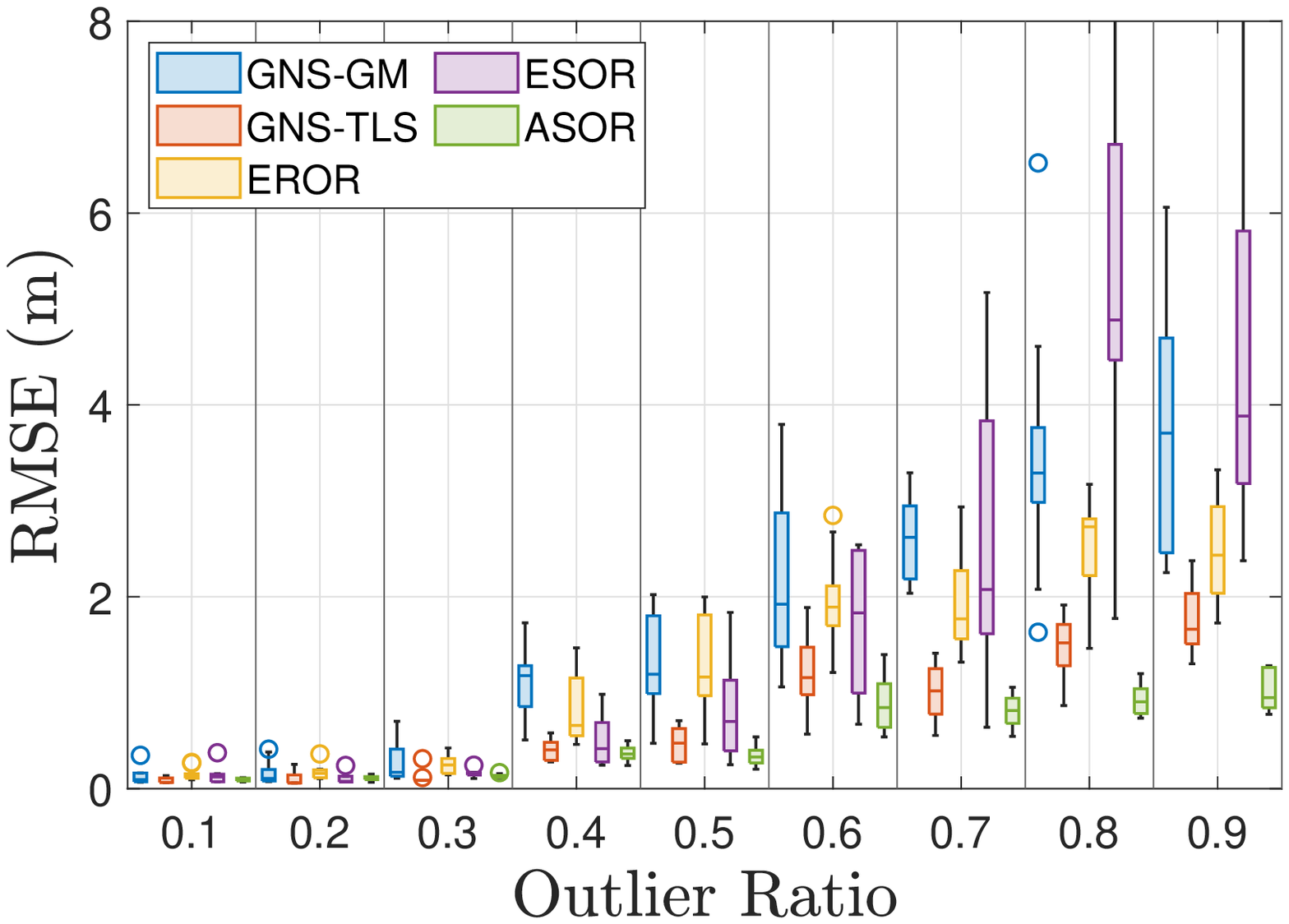}
		\caption{RMSE (INTEL).}
		\label{fig:first_PGO}
	\end{subfigure}
	\hfill
	\begin{subfigure}{0.32\textwidth}
		\includegraphics[width=\textwidth]{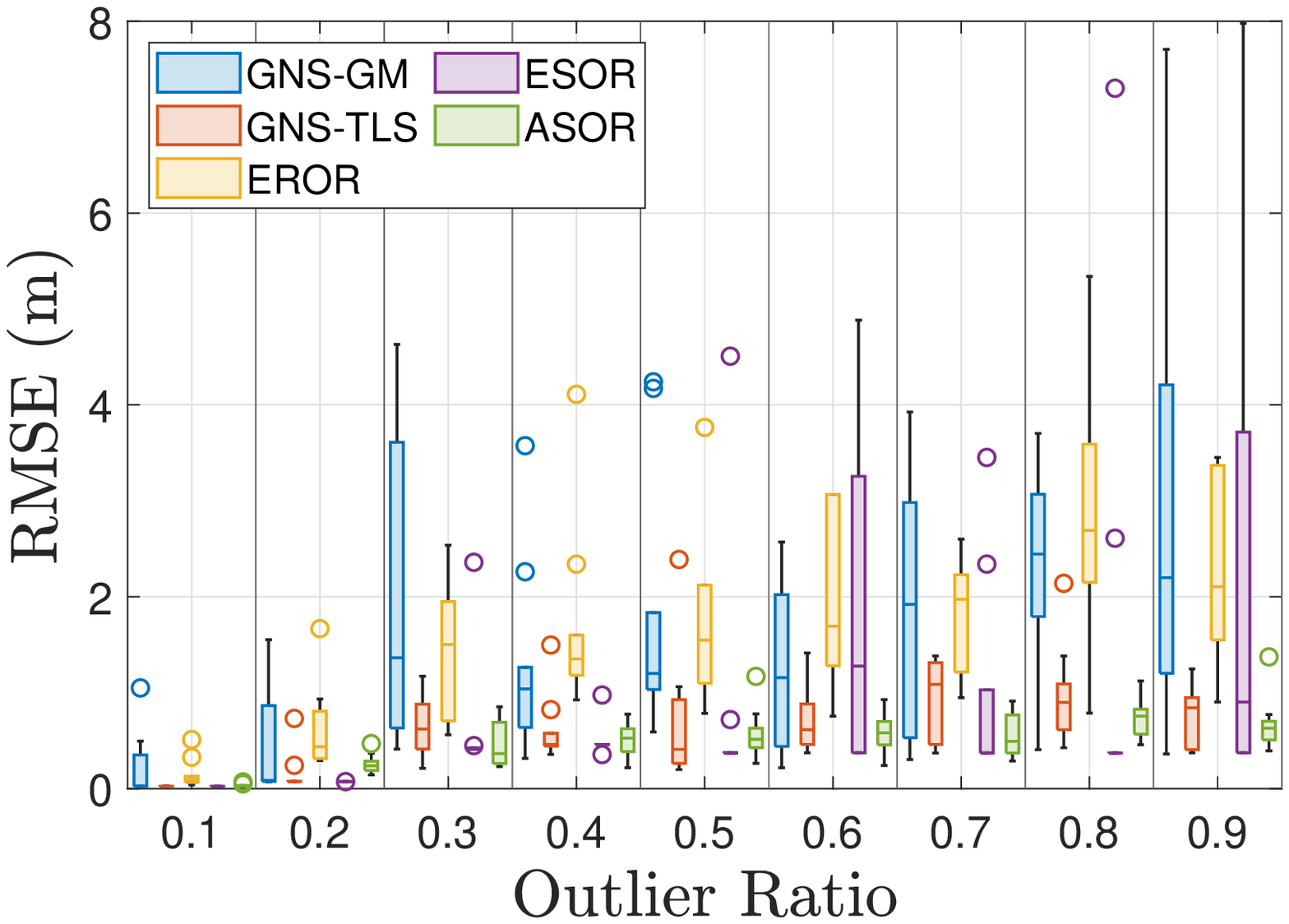}
		\caption{RMSE (CSAIL).}
		\label{fig:second_PGO}
	\end{subfigure}
	\hfill
	\begin{subfigure}{0.32\textwidth}
		\includegraphics[width=\textwidth]{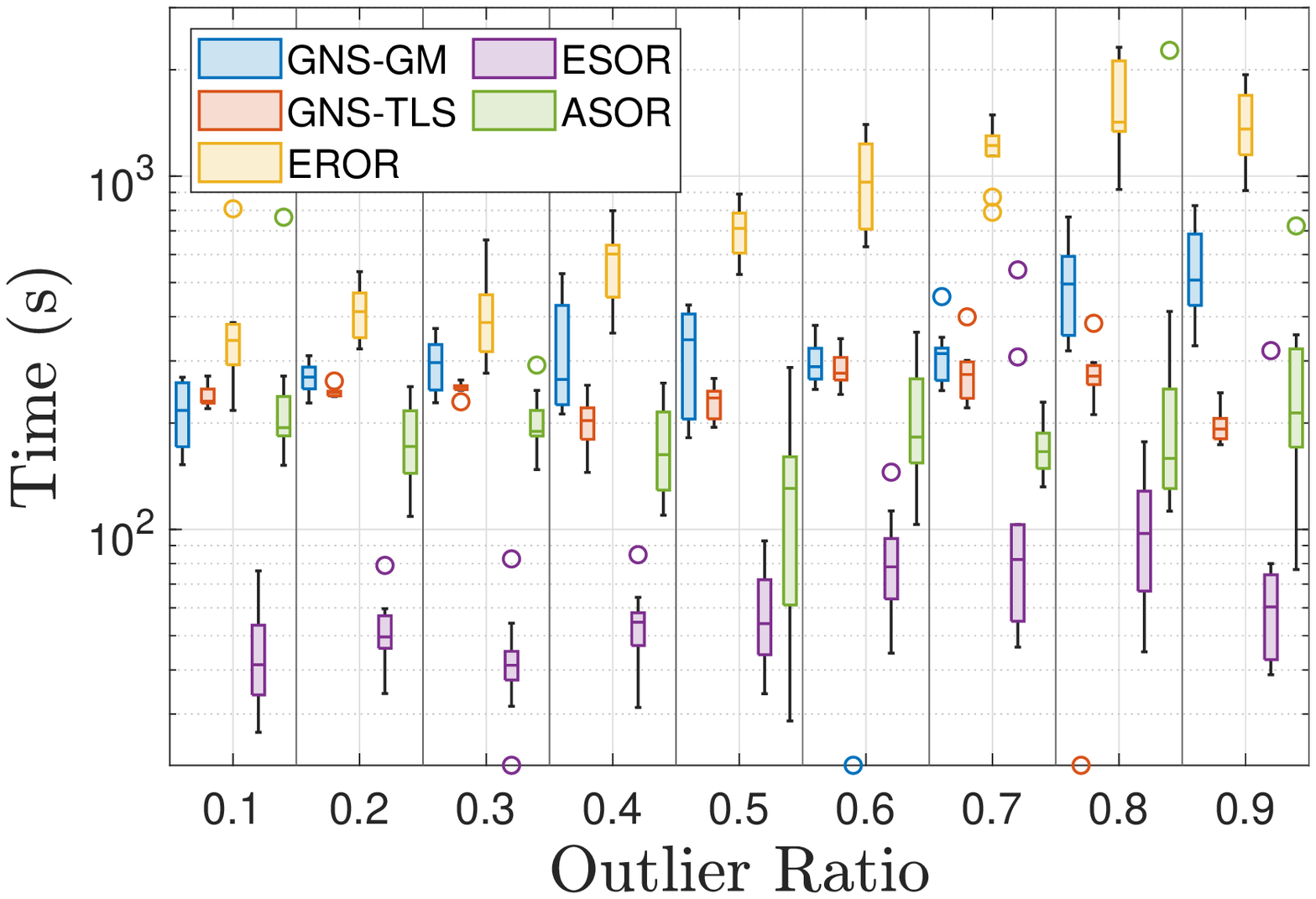}
		\caption{Computational time (INTEL).}
		\label{fig:third_PGO}
	\end{subfigure}
	\caption{Performance of robust estimators for pose graph optimization considering Intel and CSAIL datasets.}
	\label{fig:figures_PGO}
\end{figure*}  
Using the proposed EROR, ESOR and ASOR we robustify the Horn's method and report the results for the \textit{Bunny} dataset. We downsample the point cloud to $m=100$ points and restrict it within a $[-0.5, 0.5]^3$ box before applying a random rotation $\mathbf{R} \in \mathrm{SO(3)}$ and a random translation $\mathbf{t}\ (\|\mathbf{t}\|_2\leq3)$. The inliers of the transformed points are corrupted with independent noise samples drawn from $\mathcal{N}(0,0.001^2)$ whereas the outliers are randomly generated being contained within a sphere of diameter $\sqrt{3}$. 20 Monte Carlo (MC) runs are used to capture the error statistics for up to $90\%$ outlier contamination ratio.

Figs.~\ref{fig:figures_horn} showcase the performance results of our proposed methods in comparison to the GNC methods. The rotation and translational error statistics as depicted in Figs.~\ref{fig:first_horn}-\ref{fig:second_horn} are very similar in all the cases with errors generally increasing for all the methods at large outlier ratios. In terms of computational complexity, the gains of the Bayesian heuristics can be observed for this case in Fig.~\ref{fig:third_horn} which is a key evaluation parameter for runtime performance. ESOR exhibits much faster comparative behavior followed by EROR and ASOR for the entire range of outlier ratios. We have observed similar error performance of the methods for other values of $m$ (upto maximum number of available points). Also we have noticed that with an increase in the inlier noise magnitude and number of points, ASOR slows down becoming computationally comparative to the GNC methods. Moreover, we have noted that increasing the parameter $a$ generally reduces the processing time at the cost of slightly increased errors at higher outlier ratios. 

Other specialized solvers like TEASER \cite{yang2019polynomial} also exist for point cloud registration problems which are certifiably robust and can sustain higher outlier contamination better. However, it has scalability issues thanks to the involvement of sluggish semidefinite programming (SDP) based solution for rotation estimation. An improved version of TEASER namely TEASER++ \cite{9286491} was recently introduced that leverages the heuristic, GNC-TLS, in its inferential pipeline for rotation estimation while still certifying the estimates. Owing to the well-devised estimation pipeline, GNC-TLS has to deal with a smaller percentage of outliers making TEASER++ faster and improving its practicality. Since the error performance of the Bayesian heuristics is similar to the GNC methods and these are generally found to be faster in various scenarios of the point cloud registration problem this indicates their utility as standalone estimators. Moreover, these can potentially be useful in other inferential pipelines like TEASER++ (while enjoying certifiable performance) but need a thorough evaluation.

\subsection{Mesh registration}
\begin{figure}[h!]
	\centering
	\includegraphics[width=.8\linewidth,trim=1.4cm 4cm 1.4cm 3cm,clip]{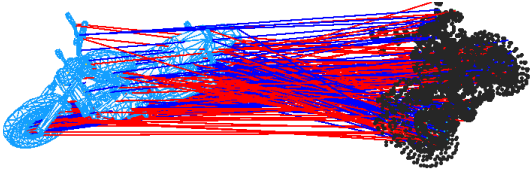}
	\caption{Mesh and point cloud with correspondences in mesh registration for the \textit{motorbike} mesh model \cite{6836101}}
	\label{fig:meshbike}
\end{figure}

In the mesh registration problem, the points $\mathbf{p}_i$ from the a point cloud are transformed to general 3D primitives $\mathbf{q}_i$ including points, lines, and/or planes. Fig.~\ref{fig:meshbike} shows the result of a random transformation of a point cloud to the \textit{motorbike} mesh model from the PASCAL dataset \cite{6836101} (blue lines: inliers, red lines: outliers). The aim is to estimate the $\mathbf{R}$ and $\mathbf{t}$ by minimizing the squared sum of residuals which represent the corresponding distances. We resort to \cite{8100078} as the basic non-minimal solver which has been proposed to find the globally optimal solution for the outlier-free case.
Using the presented EROR, ESOR and ASOR heuristics we robustify the solver and report the results for the \textit{motorbike} mesh model. To create point cloud we randomly sample points on the vertices, edges and faces of the mesh model, and then apply a random transformation $(\mathbf{R} \in \mathrm{SO(3)}$, $\mathbf{t}\ (\|\mathbf{t}{\|}_2\leq5))$ and subsequently add independent noise samples from $\mathcal{N}(0,0.01^2)$. Considering $100$ point-to-point, $80$ point-to-line and $80$ point-to-plane correspondences, we create outliers by random erroneous correspondences. For this case also, 20 MC runs are carried out to generate the error statistics for up to $90\%$ outlier contamination ratio. We see a trend similar to the point cloud registration case as depicted in Figs.~\ref{fig:first_mesh}-\ref{fig:second_mesh}. In particular, the performance in terms of errors is similar for all the algorithms. However, the Bayesian heuristics exhibit a general computational advantage, except at very high outlier ratios where the performance becomes comparable. The Bayesian heuristics generally have similar runtimes with SOR modifications having a general advantage. We observed similar performance of the methods for other combinations of correspondences in this case.

During the experiments of point cloud and mesh registration, we have noticed that ROR and SOR modifications generally get computationally more advantageous, with estimation quality remaining similar, when the outliers become larger in comparison to the nominal noise samples.

\subsection{Pose graph optimization}
\begin{figure}[h!]
	\centering
	\hspace{0cm}\subfloat[\centering {INTEL}]{{\includegraphics[width=4.2cm,trim=1cm 1cm 1cm 1cm,clip]{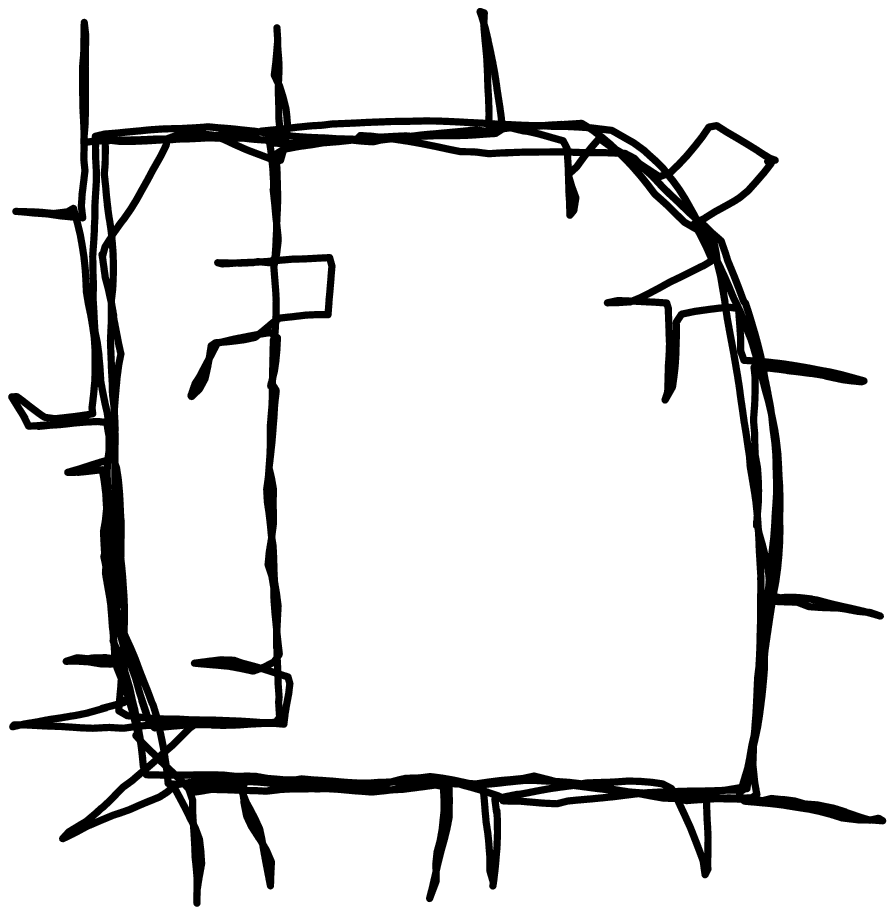} }}%
	%	\newline
	\hspace{0cm}\subfloat[\centering {CSAIL}]{{\includegraphics[width=4.2cm,trim=2cm 1cm 2cm 1cm,clip]{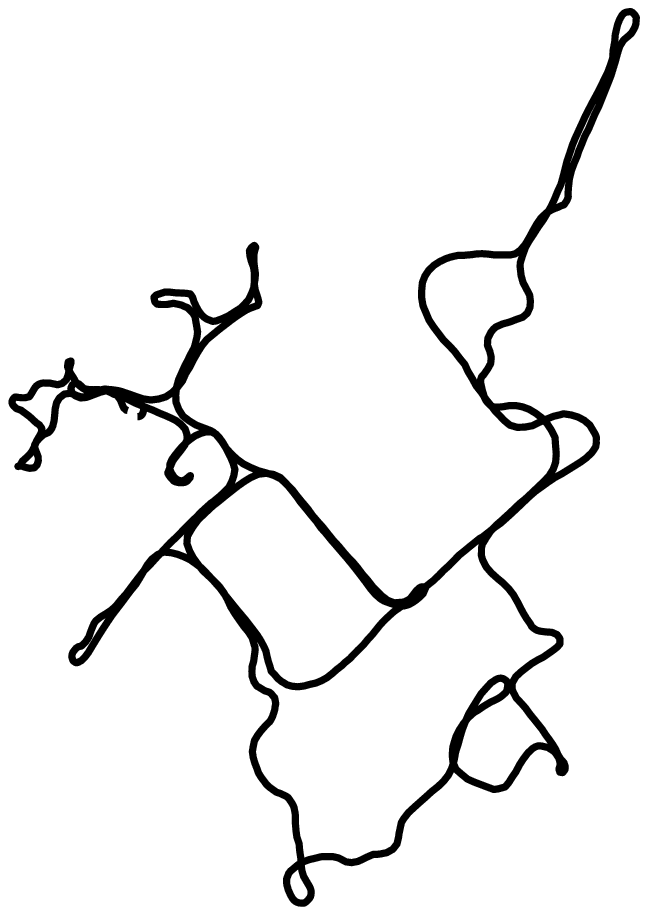} }}%
	\caption{Ground truth paths of datasets considered in pose graph optimization.}%
	\label{fig:PGO_path}%
\end{figure}

PGO is typically employed for several problems arising in robotic and computer vision applications like SLAM and structure from motion (SfM) \cite{9346012}. The objective is to estimate a set of poses $(\mathbf{t}_i,\mathbf{R}_i), i = 1, . . . , m$ using pairwise relative measurements $(\tilde{\mathbf{t}}_{i j}, \tilde{\mathbf{R}}_{i j})$. Relative observations can result in consecutive pose constraints (e.g. from odometry measurements) or non-successive pose constraints (e.g. from scan matalphang) also known as loop closures. The residual error for this case is given as 
\begin{equation}
	r\left(\mathbf{R}_i, t_i\right)=\sqrt{\kappa_{i j}\|\mathbf{R}_j-\mathbf{R}_i \tilde{\mathbf{R}}_{i j}\|_F^2+\tau_{i j}\|t_j-t_i-\mathbf{R}_i \tilde{t}_{i j}\|_2^2}\nonumber
\end{equation}
where $\kappa_{i j}$ and $\tau_{i j}$ encode the measurement noise statistics and $\|.\|_{F}$ denotes the Frobenius norm. We resort to SE-Sync \cite{rosen2019se} as the non-minimal solver for this case and use the Python binding of C++ open-sourced by the authors. Adopting the same experimentation setup of GNC we randomly corrupt loop closures and retain odometry observations. For benchmarking we consider 2D datasets including INTEL and CSAIL which are available openly (path ground truth plotted in Fig.~\ref{fig:PGO_path}). Since simulations take much more time as compared to the previous case we carry out 10 MC runs to obtain the error statistics. Fig.~\ref{fig:first_PGO} showcases the root mean squared (RMSE) of the trajectory considering the INTEL dataset. The proposed Bayesian heuristics in this case also exhibit comparative performance to the GNC methods. At very higher outlier rates ESOR has slightly compromised performance. In Fig.~\ref{fig:second_PGO} the RMSE for the CSAIL dataset are depicted reflecting the same pattern. In the PGO examples, GNC-TLS and ASOR outperform other methods with ASOR generally having the least error for different outlier ratios. As far as the computational performance is concerned ESOR has the smallest runtime, followed by ASOR while ROR is the slowest for both cases. Fig.~\refeq{fig:third_PGO}  depicts the computational runtime statistics for the INTEL dataset. CSAIL has similar computational time statistics with ESOR being relatively much faster as compared to other methods.     

%
%\begin{figure*}[h!]
%	\centering
%	\begin{subfigure}{0.32\textwidth}
	%		\includegraphics[width=\textwidth]{CSAIL_rmse.eps}
	%		\caption{CSAIL subfigure.}
	%		\label{fig:first}
	%	\end{subfigure}
%	\hfill
%	\begin{subfigure}{0.32\textwidth}
	%		\includegraphics[width=\textwidth]{CSAIL_time.eps}
	%		\caption{CSAILsubfigure.}
	%		\label{fig:second}
	%	\end{subfigure}
%	\caption{Creating subfigures in \LaTeX.}
%	\label{fig:figures}
%\end{figure*}  
%
%
%\begin{figure*}[h!]
%	\centering
%	\begin{subfigure}{0.32\textwidth}
	%		\includegraphics[width=\textwidth]{intel_rmse.eps}
	%		\caption{Intel subfigure.}
	%		\label{fig:first}
	%	\end{subfigure}
%	\hfill
%	\begin{subfigure}{0.32\textwidth}
	%		\includegraphics[width=\textwidth]{intel_time.eps}
	%		\caption{Intel subfigure.}
	%		\label{fig:second}
	%	\end{subfigure}
%	\caption{Creating subfigures in \LaTeX.}
%	\label{fig:figures}
%\end{figure*}  
Lastly, we also evaluated the Bayesian heuristics using $\max_i({w^{i} {\hat{r}_i^{2}}})<\bar{c}^2$ as the stopping criteria which resulted in much faster performance but with slightly degraded error performance in the considered scenarios of the spatial perception applications. {Supplementary results with other experimental data and the source code is openly available \cite{git_ref}}.

\section{Conclusion}\label{Sec_conc}
We have proposed three Bayesian heuristics: EROR, ESOR and ASOR as nonlinear estimators for spatial perception problems. Like the existing general-purpose GNC heuristics, these have the ability to invoke existing non-minimal solvers. Evaluations in several experiments demonstrate their merits as compared to the GNC heuristics. In particular, in the 3D point cloud and mesh registration problems EROR, ESOR and ASOR have similar estimation errors over a wide range of outlier ratios. However, the Bayesian heuristics have a general advantage in computational terms. For the PGO setups, we generally find the Bayesian methods to compete with GNC in estimation quality. The devised ROR and SOR modifications are found to be the least and most computationally efficient for this case. Using another suggested criteria can lead to further improvement in computational terms at expense of estimation quality. In short, the proposed methods provide general purpose options, in addition to the GNC heuristics, to robustify the existing non-minimal solvers against outliers in different spatial perception applications indicating their usefulness. Empirical evidence suggests that the proposed approaches provide a general edge in computational terms while remaining competitive in terms of error. The actual possible gains depend on whether the solvers are used standalone or in an inferential pipeline for the particular application scenarios and should be evaluated for the case under consideration before deployment. We believe that the work can be further extended in different directions by aiming to devise the heuristics without knowledge of nominal noise statistics. Moreover, these methods can also be tested within hybrid approaches where the estimates subsequently get certified for optimality.

\bibliographystyle{IEEEtran}
\bibliography{bibli}

\end{document}